\documentclass[review]{elsarticle}

\usepackage{amssymb}
\usepackage{amsmath}
\usepackage[ruled]{algorithm2e}
\usepackage{algorithmic}
\usepackage{array}
\usepackage{booktabs}
\usepackage{multirow}
\usepackage{subfigure}
\usepackage{lineno}
\usepackage{hyperref}
\usepackage{color}

\begin{document}

\begin{frontmatter}



\title{Self-Attention Empowered Graph Convolutional Network for Structure Learning and Node Embedding}


\author[mymainaddress]{Mengying Jiang}
\ead{myjiang@stu.xjtu.edu.cn}

\author[mymainaddress]{Guizhong Liu\corref{mycorrespondingauthor}}
\cortext[mycorrespondingauthor]{Corresponding author}
\ead{liugz@xjtu.edu.cn}

\author[mysecondaryaddress]{Yuanchao Su}
\ead{suych3@xust.edu.cn}

\author[mymainaddress]{Xinliang Wu}
\ead{wuliang@stu.xjtu.edu.cn}

\address[mymainaddress]{School of Electronic and Information Engineering, Xi'an Jiaotong University, Xi'an 710049, China}
\address[mysecondaryaddress]{College of Geomatics, Xi'an University of Science and Technology, Xi'an 710054, China}

\begin{abstract}
In representation learning on graph-structured data, many popular graph neural networks (GNNs) fail to capture long-range dependencies, leading to performance degradation. Furthermore, this weakness is magnified when the concerned graph is characterized by heterophily (low homophily). To solve this issue, this paper proposes a novel graph learning framework called the graph convolutional network with self-attention (GCN-SA). The proposed scheme exhibits an exceptional generalization capability in node-level representation learning. The proposed GCN-SA contains two enhancements corresponding to edges and node features. 
For edges, we utilize a self-attention mechanism to design a stable and effective graph-structure-learning module that can capture the internal correlation between any pair of nodes.
This graph-structure-learning module can identify reliable neighbors for each node from the entire graph.
Regarding the node features, we modify the transformer block to make it more applicable to enable GCN to fuse valuable information from the entire graph. 
These two enhancements work in distinct ways to help our GCN-SA capture long-range dependencies, enabling it to perform representation learning on graphs with varying levels of homophily.
The experimental results on benchmark datasets demonstrate the effectiveness of the proposed GCN-SA. Compared to other outstanding GNN counterparts, the proposed GCN-SA is competitive. The source code is available at \href{https://github.com/mengyingjiang/GCN-SA}{https://github.com/mengyingjiang/GCN-SA.}

\end{abstract}



\begin{keyword}
Representation learning; Heterophily; Structure learning; Graph neural networks

\end{keyword}

\end{frontmatter}



\section{Introduction} \label{sect:review}
Graph neural networks (GNNs) are capable of graph representation learning with many applications ranging from the Internet of things to knowledge graphs \cite{{PR1}}. 
Many GNNs have been developed in recent years, including graph attention network (GAT)~\cite{GAT+chapter2018}, graph convolutional network (GCN)~\cite{GCN+chapter2017}, graph isomorphism network (GIN)~\cite{GIN+chapter2019}, and GraphSAGE~\cite{GraphSage+chapter2017}.
For GNNs, each node iteratively updates and improves its feature representation by aggregating the ones of its neighbors and the node itself~\cite{{neighbor}}.
Typically, neighbors are defined as the set of one-hop neighbors in a graph~\cite{PR9}.
A variety of aggregation functions have been adapted to GNNs, such as mean, weighted sum, and maximum.~\cite{{pna+chapter2020}}.

In recent years, convolutional neural networks (CNNs) have grown substantially and have achieved significant success in a wide range of tasks~\cite{{ccli1}}.
GCNs are generalizations of classical CNNs that handle graph data, such as
point clouds, molecular data, and social networks~\cite{Fastgcn}.
As the most attractive GNNs, GCNs have been widely used in various applications~\cite{{hamilton+chapter2017}}.
Nevertheless, GCNs may not capture long-range dependencies in graphs because GCNs update feature representations by simply summing the normalized feature representations from all adjacent nodes~\cite{{PR3}}.
This fundamental limitation significantly limits the ability of GCNs to represent graph-structured data~\cite{NLGNN}.
Furthermore, this weakness is magnified in graphs with low or medium levels of homophily~\cite{PR4}.

Homophily is an essential characteristic of graph-structured data~\cite{Reviewer51}, in which connected nodes often share similar characteristics and possess the same labels~\cite{Reviewer52}.
For instance, friends tend to have similar interests or ages, and a scientific paper and its citations are usually from the same research area~\cite{{PR10}}.
However, in the real world, there are also settings about ``opposites attract,'' leading to graphs with heterophily (or low/medium level of homophily)~\cite{rlgjkn}. In such cases, the proximal nodes often belong to different classes and exhibit dissimilar features~\cite{{Defending}}.
For example, in terms of device type, there are many classes such as Smart Fitness, Car, Smartphone, and the social Internet-of-Things (SIOT) system~\cite{Things}.
Devices (nodes) often interact with different types of devices in SIOT graphs.
Most existing GNNs assume graphs are under high-level homophily, including GAT, GCN~\cite{hhgcn}.
Consequently, many GNNs perform poorly in generalizing graphs with low/medium homophily, even worse than the multi-layer perception (MLP)~\cite{MLP}, which relies solely on node features~\cite{fraudsters}.

Recently, researchers have developed new approaches to solve this problem~\cite{PR8}, 
such as geometric GCN (Geom-GCN)~\cite{Geom+chapter2020}, heterophily and homophily GCN (H$_{2}$GCN)~\cite{hhgcn}, and iterative deep graph learning (IDGL)~\cite{IDGL}.
Geom-GCN proposes a geometric aggregation scheme to learn structural information, which enables GCNs to achieve an enhanced learning performance~\cite{Geom+chapter2020}.
However, Geom-GCN only focuses on local neighborhood structures and overlooks cases in which neighborhood nodes may not provide valuable information in graphs with low/medium homophily.
Thus, the classification performance of Geom-GCN is typically unsatisfactory when the graphs are heterophilic.
H$_{2}$GCN improves the classification performance of GCN via separating ego and neighbor-embeddings~\cite{hhgcn}.
However, each node only aggregates feature representations from adjacent nodes and itself, leading H$_{2}$GCN cannot capture long-range dependencies.
IDGL~\cite{IDGL} achieved a good classification performance by jointly and iteratively learning node embeddings and a new graph structure, whereas the iterative update of the optimized graph structure was time-consuming.
In addition, IDGL includes many hyper-parameters that need to be tuned in advance, and the values of the hyper-parameters vary significantly between different graph datasets,
which limits the generalization ability of IDGL.

This study proposes a novel graph learning framework called a graph convolutional network with self-attention (GCN-SA) to address the abovementioned issues.
The proposed GCN-SA introduces two key improvements, one pertaining to edges and the other to nodes.
From the perspective of edges, we generate an optimized re-connected adjacency matrix by jointly learning the graph structure and node embeddings.
Inspired by the ability of the self-attention mechanism to reduce dependence on external information and capture the internal correlation of nodes~\cite{self-atten},
we employ the multi-head self-attention (MHSA) mechanism to learn a new adjacency matrix. Multi-head attention helps to stabilize the learning process of self-attention~\cite{adaptivegcn+chapter2018}.
Specifically, we use the MHSA mechanism to calculate the attention scores between nodes. 
Then, we jointly apply the k-nearest neighbors (KNN) and minimum-threshold methods to select the nodes with the highest attention scores as the new adjacent nodes of the target node.
Subsequently, the new graph structure and node embeddings are jointly optimized for the downstream prediction task.
In summary, the structure-learning module can identify reliable neighbors for each node from the entire graph to help the model capture long-range dependencies through an MHSA mechanism, an appropriate screening mechanism, and the joint optimization of the node embeddings and graph structure.
Although IDGL~\cite{IDGL} can also learn an optimized graph structure through structural learning, the optimized adjacency matrix in IDGL is an adjustment of the original graph structure.
The re-connected graph structure in our GCN-SA is related to the internal correlation of the nodes.
Moreover, the re-connected graph structure does not replace the original one but instead works together with the original one to perform the downstream task.
For the proposed GCN-SA, cooperation enables both the re-connected and original graph structures to play their respective strengths and adapt graphs with various homophily levels.
From the perspective of node features, we modify the transformer block to make it more applicable to GCN.
We then use the modified transformer block to perform node feature fusion and embedding fusion to integrate valuable information from the entire graph.
Each transformer block with a self-attention mechanism includes two sub-layers: an MHSA mechanism and a fully connected feed-forward network.
In addition, each sub-layer is followed by a residual connection and layer normalization.
Nevertheless, GCN is sensitive to over-fitting.
Therefore, we modify the transformer block to fuse valuable information while avoiding over-fitting.
Specifically, we retain the MHSA mechanism to preserve the ability to capture the internal correlation of nodes.
Subsequently, we apply two dropouts and two residual connections after applying the MHSA mechanism.
The modified transformer block is utilized for node feature fusion. The fusional feature vectors are then combined with the original feature vectors to form node ego-embeddings.
Afterward, we apply the original and re-connected adjacency matrices to perform feature aggregation on the ego-embeddings, generating neighbor-embeddings and reconnected-neighbor-embeddings, respectively.
Next, we concatenate ego-embeddings, neighbor-embeddings, and reconnected-neighbor-embeddings to form representative node embeddings. 
Subsequently, We reuse the modified transformer block to fuse valuable node embeddings from the entire graph.
Finally, the fusional result is set as the final node representation for node classification.

Compared with other GNNs, the contributions of the proposed GCN-SA can be summarized as follows:

1) The proposed GCN-SA introduces a new learning framework to build the reconnected adjacency matrix using an MHSA mechanism.

2) Our modified transformer block enables the GCN to integrate valuable information from an entire graph effectively. 

3) We concatenate the ego-embeddings, neighbor-embeddings, and reconnected-neighbor-embeddings so that they could play to their respective strengths to adapt to graphs with various homophily levels.

4) The proposed GCN-SA is the first to improve the GNN from both edges and node features by utilizing a self-attention mechanism. The self-attention mechanism can construct a fully connected graph, allowing the model to capture the internal associations between any pair of nodes. This capability enables our GCN-SA to capture long-range dependencies effectively.

\begin{table}[htp]\footnotesize
\caption{Commonly used notations}
\label{def}
\begin{center}
\vspace{-3mm}
\setlength\tabcolsep{1pt}
\begin{tabular}{c |c}
\toprule
Notations& Descriptions\\ \hline  
$\mathcal{G}$ & A graph.\\\hline
$\odot$  & Hadamard product. \\\hline
$\|$  & Concatenation. \\\hline
$\mathcal{V}$ & The set of edges in a graph. \\\hline
$v_{i}$ & A node $v_{i} \in \mathcal{V}$.\\\hline
$n$& The number of nodes, $n=|\mathcal{V}|$.\\\hline
$c$& The number of node labels.\\\hline
$m$& The number of attention heads.\\\hline
$d$& The dimension of a node feature vector.\\\hline 
$\mathcal{N}_{i}$& The set of neighbors for node $v_{i}$. \\\hline
${\bf A}$ &The adjacency matrix of the graph. \\\hline
${\bf D}$ & The degree matrix of ${\bf A}$, $D_{ii}=\sum_{j} A_{i, j}$.\\\hline
$\hat{{\bf A}}$ & The normalized ${\bf A}$ with self-loop.\\\hline
${\bf A}_{*}$ &The re-connected adjacency matrix. \\\hline
$\hat{{\bf A}}_{*}$ &The normalized ${\bf A}_{*}$\\\hline
${\bf S}$ & The attention score matrix of a graph.\\\hline
${\bf X} \in \mathbb{R}^{n \times d}$ &The node feature matrix.\\\hline
${\bf x}_{i} \in \mathbb{R}^{d}$& The feature vector of the node $v_{i}$.\\\hline 
${\bf H}$ &The ego-embeddings of nodes.\\\hline
${\bf H}_{{\bf A}_{*}}$ &The reconnected-neighbor-embeddings of nodes.\\\hline
${\bf H}_{{\bf A}}^{(K)}$ &The neighbor-embeddings of nodes.\\\hline
${\bf H}^{cb}$ &The general embeddings of nodes.\\\hline
${\bf Z} \in \mathbb{R}^{n \times c}$& The probability distributions of nodes.\\\hline
${\bf W}^{0}$, ${\bf W}^{1}$, ${\bf W}_{i}^{Q}$&\multirow{2}{*}{Learnable model parameters}\\
${\bf W}_{i}^{K}$, ${\bf W}_{i}^{V}$, ${\bf w}$&\\
\bottomrule
\end{tabular}
\end{center}
\end{table}

\section{Preliminaries}
\label{related work}

This study proposes a novel GNN model related to GCN and transformer~\cite{self-atten}.
The minimal set of definitions is presented in Subsection~\ref{definition}.
The theories about the GCN and transformer are reviewed in Subsection~\ref{GCN} and~\ref{Transformer}, respectively.

\subsection{Definition}
\label{definition}

Throughout this paper, lowercase characters represent scalars, and bold lowercase characters and bold uppercase characters represent vectors and matrices, respectively.
Unless otherwise specified, the notations employed in this study are listed in Table~\ref{def}.

Let $\mathcal{G}=(\mathcal{V}, \mathcal{E})$ be an undirected graph, where $\mathcal{V}$ and $\mathcal{E}$ represent the sets of nodes and edges, respectively. 
$|\mathcal{V}|=n$ is the number of nodes and 
$v_{i} \in \mathcal{V}$ denotes $i$-th node.
$e_{ij}\in \mathcal{E}$ represents the edge between $v_{i}$ and $v_{j}$.
$\mathcal{N}_{i}=\left\{v_{j}\in \mathcal{V} | e_{ij}\in \mathcal{E}\right\}$ represents the set of one-hop neighbors for node $i$.
${\bf A}\in \mathbb{R}^{n \times n}$ is the adjacency matrix.
$A_{ij}=0$ if $e_{ij} \notin \mathcal{E}$ and $A_{ij}=1$ if $e_{ij} \in \mathcal{E}$.
Let ${\bf X}= \left\{{\bf x}_i\right\}_{i=1}^N \in \mathbb{R}^{n \times d}$ be the feature matrix for nodes, where ${\bf x}_{i}\in \mathbb{R}^{1 \times d}$ denotes the feature vector of node $v_{i}$.
Each node is associated with a label.
Given a multilayered
network and the semantic labels ${\bf y}_{lab}$ for a subset of nodes $\mathcal{V}_{lab} \in \mathcal{V}$, where $y \in {\bf y}_{lab}$ means one
of the $c$ predefined classes.
GNNs can learn the feature representations of nodes and graphs by using the graph structure and node features.
Then the task of node classification is to predict the label for each node without label $v_{i}\in \mathcal{V} |v_{i} \notin \mathcal{V}_{lab}$ according to the feature representations of the corresponding node~\cite{Comprehensive}.

\subsection{GCN}
\label{GCN}
Following the work in~\cite{GCN+chapter2017}, GCN updates node features by adopting an isotropic averaging operation over the feature representations of one-hop neighbors.
Let ${\bf h}_{j}^{\ell}$ be the feature representation of node $v_{j}$ in the $l$-th GCN layer, 
then a single message passing step in the GCN model takes the following form:
\begin{equation}
{\bf h}_{i}^{\ell+1} =  {\rm ReLU}\left(\sum_{j \in \mathcal{N}_{i}} \frac{1}{\sqrt{{\rm deg}_{i}{\rm deg}_{j}}} {\bf h}_{j}^{\ell} {\bf W}^{\ell}\right),
\label{eq:GCNlayer}
\end{equation}
where ${\bf W}^{\ell}$ is a learnable weight matrix for the $l$-th GCN layer, $\mathcal{N}_{i}$ represents the set of one-hop neighbors of node $v_{i}$.
Note that ${\rm deg}_{i}$ and ${\rm deg}_{j}$ denote the in-degrees of nodes $v_{i}$ and $v_{j}$, respectively. $\operatorname{ReLU}(\cdot)=\max (0, \cdot)$ is the nonlinear activation function.
Along this line, the forward model of a 2-layer GCN~\cite{GCN+chapter2017} can be represented as:
\begin{equation}
{\bf Z}=\displaystyle f({\bf X}, {\bf A})=\displaystyle softmax\left(\hat{{\bf A}} {\rm ReLU}\left(\hat{{\bf A}} {\bf X} {\bf W}^{0}\right) {\bf W}^{1}\right),
\label{eq:2layerGCN}
\end{equation}
where ${\bf A}\in \mathbb{R}^{n \times n}$ represents the adjacency matrix, and ${\bf X}\in \mathbb{R}^{n \times d}$ is the feature matrix of nodes and is also the input of the first GCN layer.
$n$ is the number of nodes, and $d$ is the feature dimension of nodes.
$\hat{{\bf A}}=\widetilde{{\bf D}}^{-1 / 2} \tilde{{\bf A}} \widetilde{{\bf D}}^{-1 / 2}$ represents the normalized $\tilde{{\bf A}}$, where $\tilde{{\bf A}} = {\bf A}+{\bf I}$ represents ${\bf A}$ with self-loop, and ${\bf I}\in \mathbb{R}^{n \times n}$ is an identity matrix.
$\tilde{{\bf D}}$ is the degree matrix of $\tilde{{\bf A}}$.
Each element $\hat{A}_{i, j}$ is defined as:
\begin{equation}
\hat{A}_{i, j}=\left\{\begin{array}{cl}
\frac{1}{\sqrt{{\rm deg}_{i}{\rm deg}_{j}}} &{\rm nodes~} i, j {\rm ~are~one-hop~neighbors}, \\
0 & {\rm otherwise }.
\end{array}\right.
\label{eq:Aij}
\end{equation}

\subsection{Transformer Block}
\label{Transformer}
The SA mechanism can learn the weight distributions of different features~\cite{PR11}. In other words, the SA mechanism can help the network to recognize features that are important for prediction. 
A transformer block consists of two sub-layers: an MHSA mechanism is the first sub-layer, and a simple fully connected feed-forward network is the second sub-layer~\cite{self-atten}.
The MHSA mechanism is calculated as follows:
\begin{equation}
{\bf X}_\text{MHSA}=\text{Concat}\left(\text {head}_{1},\text{head}_{2},\ldots,\text{head}_{\mathrm{m}}\right){\bf W}^{0},
\label{concat0}
\end{equation}
\begin{equation}
\text { head }_{i}=\operatorname{softmax}\left(\frac{{\bf Q}_{i} {\bf K}_{i}^{\mathrm{T}}}{\sqrt{d_{k}}}\right) {\bf V}_{i},
\label{innerproduct0}
\end{equation}
\begin{equation}
{\bf Q}_{i}={\bf X} {\bf W}_{i}^{Q},
\label{Q0}
\end{equation}
\begin{equation}
{\bf K}_{i}={\bf X}{\bf W}_{i}^{K},
\label{K0}
\end{equation}
\begin{equation}
{\bf V}_{i}={\bf X} {\bf W}_{i}^{V},
\label{V0}
\end{equation}
where ${\bf X}$ is the input. ${\bf Q}_{i}$, ${\bf K}_{i}$, and ${\bf V}_{i} $ respectively correspond to the Query, Key, and Value of the $i$-th head ($i=1,\dots,m$). 
Moreover, a residual connection~\cite{resnet} and layer normalization~\cite{layernorm} follow each sub-layer. Thus, the input of the second sub-layer is as follows:
\begin{equation}
{\bf X}_\text{sublayer1}= \operatorname{LayerNorm} ({\bf X}+ {\bf X}_\text{MHSA}).
\label{RESNET0}
\end{equation}
The second sub-layer is a fully connected feed-forward
network. Thus, the output of the transformer block is calculated
as follows:
\begin{equation}
{\bf X}_\text{sublayer2}= \operatorname{LayerNorm} ({\bf X}_\text{sublayer1}+ {\bf X}_\text{sublayer1}{\bf W}^{1}).
\label{RESNET1}
\end{equation}

\begin{figure*}[t]
\centering
\includegraphics[width=4.8in,height=1.85in]{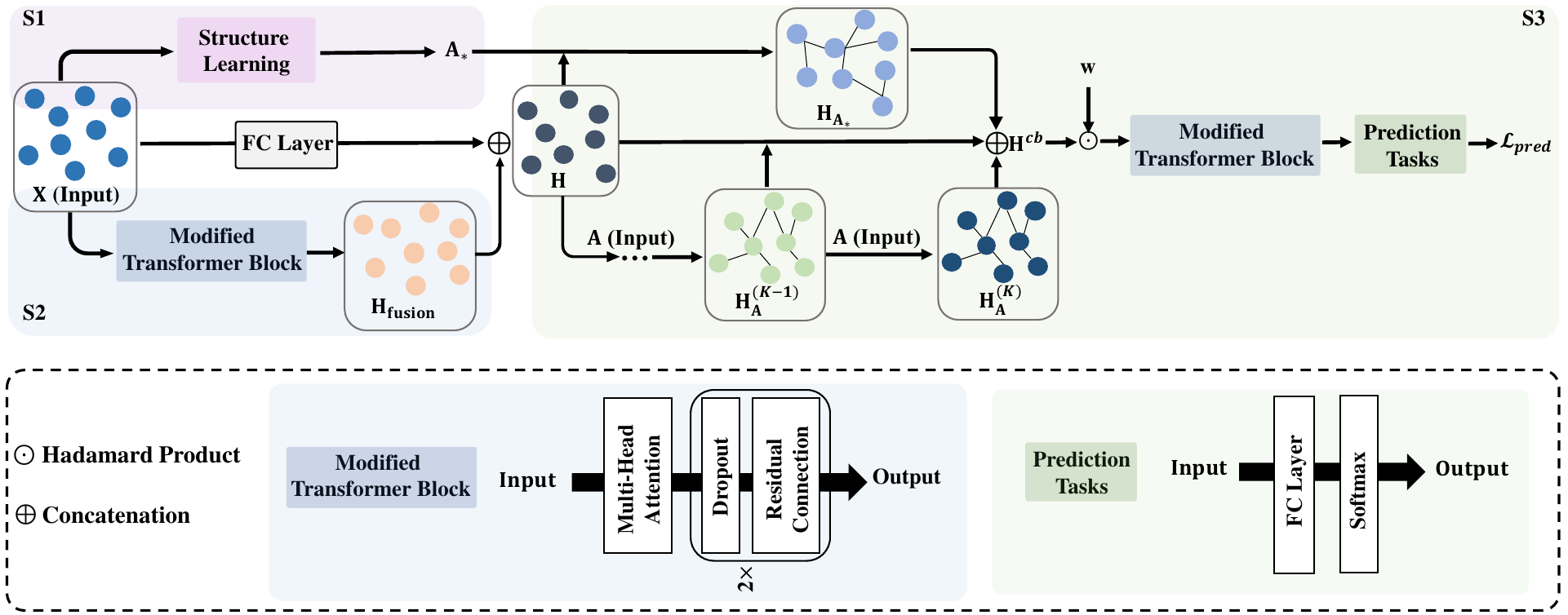}
\vspace{-0.3cm}
\caption{GCN-SA consists of three stages:
(S1) \textit{re-connected graph ${\bf A}_{*}$ learning}, (S2) \textit{fusional feature learning}, and (S3) \textit{graph convolutional network with self-attention (GCN-SA)}.
In (S1), we construct a re-connected adjacency matrix ${\bf A}_{*}$ through attention score learning.
This process allows the reconnected graph to be gradually optimized with the evolution of node embeddings.
In (S2), we design and employ a modified transformer block to perform feature vector fusion.
In (S3), we combine the original and fusional feature vectors as the ego-embeddings ${\bf H}$.
Then we perform feature aggregation on ${\bf H}$ using ${\bf A}_{*}$ and ${\bf A}$, respectively.
Subsequently, the ${\bf H}$ and the results of feature aggregation are concatenated as the ${\bf H}^{cb}$, and we use a learnable weighted vector ${\bf w}$ to highlight the crucial dimensions of ${\bf H}^{cb}$.
Finally, we reuse the modified transformer block to perform node embedding fusion.}
\label{fig:Fig.1}
\end{figure*}

\section{GCN-SA}
\label{headings}
This section presents a novel GNN, the graph convolutional network with self-attention (GCN-SA), for node classification of graph-structured data. Figure~\ref{fig:Fig.1} shows the pipeline of our GCN-SA.
The remainder of this Section is organized as follows.
Subsection~\ref{adjacent} gives the construction details of the re-connected adjacency matrix.
Subsection~\ref{AM} describes the modified transformer block and the generation of fusional feature vectors.
Subsection~\ref{gcn-ef} elaborates on the proposed GCN-SA.

\subsection{Re-connected Graph}\label{adjacent}
\label{Re-connected}

Most existing GNNs are designed for graphs with high homophily, where the connected nodes usually possess similar feature representations and have the same label, for example, citation and community networks~\cite{{Jointly}}.
However, there are many graphs with low or medium homophily in the real world, where the connected nodes often have distinct features and possess different labels, such as webpage linking networks~\cite{{struc2vec}}.
In summary, in practical applications, GNNs designed under the assumption of high homophily are inappropriate for graphs with low or medium homophily.

To alleviate this problem, we build a re-connected adjacency matrix via structure learning to provide each node with reliable neighbor information.
Specifically, we use the MHSA mechanism to compute the attention scores between nodes to explore their internal correlation.
Considering our primary focus on undirected graphs, generating undirected reconstructed graphs aligns more closely with our practical requirements.
Therefore, the MHSA mechanism is modified as follows:
\begin{equation}
{\bf S}=\text{Mean}\left(\text {head}_{1},\text{head}_{2},\ldots,\text{head}_{\mathrm{m}}\right),
\label{eq:MH1}
\end{equation}
\begin{equation}
\text {head}_{i}=\operatorname{cosine}\left({{\bf Q}_{i}, {\bf Q}_{i}}\right),
\label{eq:MH2}
\end{equation}
\begin{equation}
{\bf Q}_{i}= {\bf X} {\bf W}_{i}^{Q}, 
\label{eq:MH3}
\end{equation} 
where ${\bf X}\in \mathbb{R}^{n \times d}$ represents the node feature matrix. ${\bf W}_{i}^{Q}\in \mathbb{R}^{d\times p}$ denotes the learnable weighted matrix for the $i$-th head ($i = 1, \dots, m$).
Eq.~\ref{eq:MH2} calculates the cosine scores between each pair of row vectors in matrix ${\bf Q}_{i}$, resulting in the cosine score matrix, $\text{head}_{i}\in \mathbb{R}^{n \times n}$.
In this manner, we can obtain a symmetric attention score matrix ${\bf S}\in \mathbb{R}^{n \times n}$, where the element $S_{ij}$ ranges between $[-1,1]$.
The MHSA mechanism can construct multiple fully connected graphs, enabling each node to connect to all the other nodes. 
This capability allows each node to aggregate features from any other node, but it may result in excessive computational burden and introduce noise. 
However, the re-connected adjacency matrix is expected to be sparse, connected, and non-negative~\cite{IDGL}. 
To address this issue, we combine the KNN and minimum-threshold methods to select the most reliable neighbors for each node.
Specifically, we first define a positive integer $r$ and a non-negative threshold $\epsilon$.
Next, we retain attention scores that are either greater than $\epsilon$ or belong to the set of $r$ largest elements in the corresponding row while setting the remaining attention scores to zero.
The new adjacency matrix ${\bf A}_{*}\in \mathbb{R}^{n\times n}$ is represented as:
\begin{equation}
A_{*ij}=\left\{\begin{array}{cl}S_{ij} & S_{ij}>\epsilon~or~S_{ij} \in \mathcal{R}_{i},\\
0 & {\rm otherwise },
\end{array}\right.
\label{S}
\end{equation}
where $\mathcal{R}$ denotes the the set of $r$ largest elements in the $i$-$th$ row of ${\bf S}$.
To guarantee the symmetry of ${\bf A}_{*}$, we impose the following constraint on ${\bf A}_{*}$:
\begin{equation}
A_{*ij}= \operatorname{max}(A_{*ij},A_{*ji}).
\label{S1}
\end{equation}

\begin{figure*}[t]
\centering
\includegraphics[width=4in,height=3.8in]{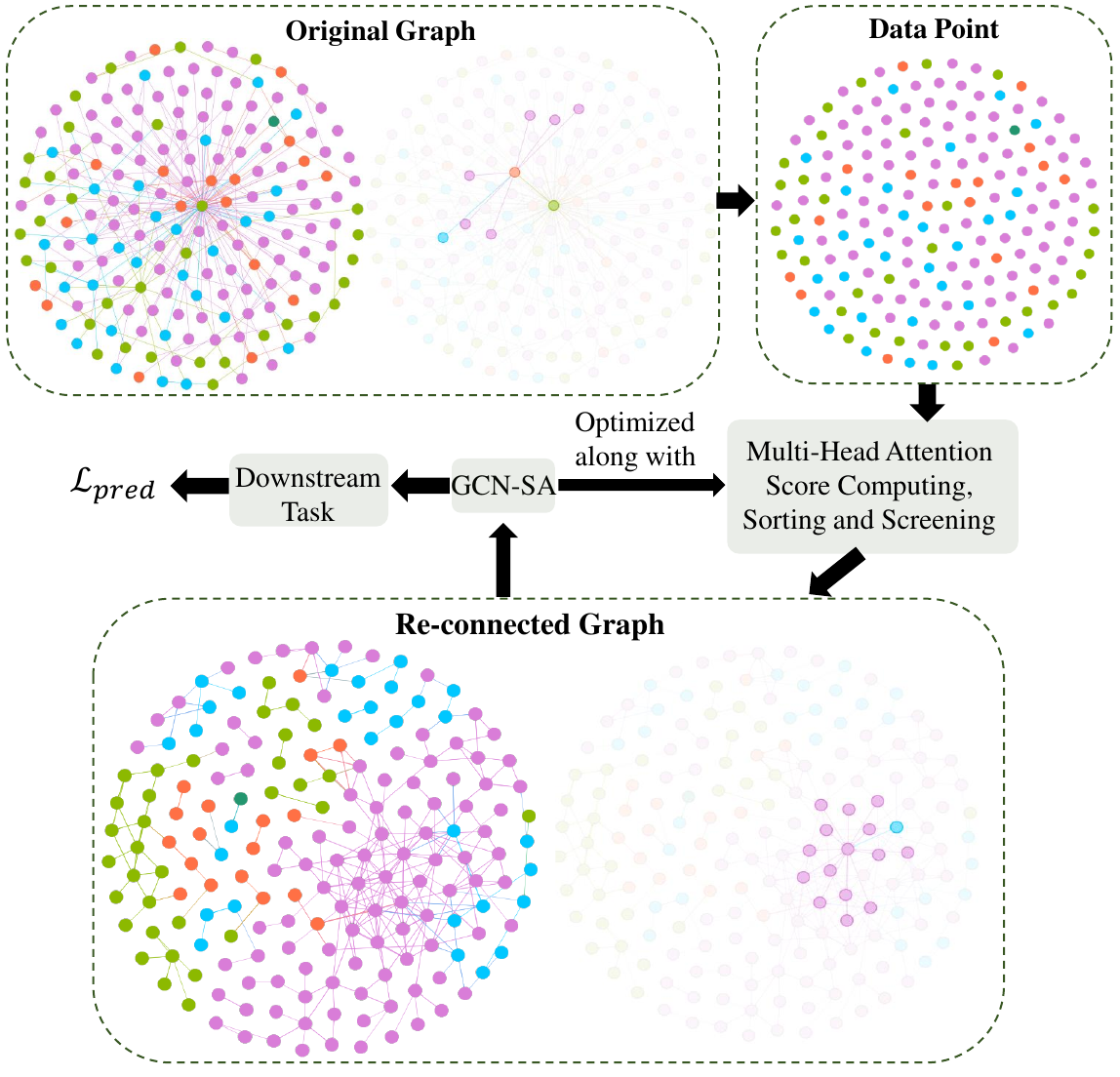}
\vspace{-5mm}
\caption{Flowchart of the re-connected adjacency matrix learning. The graph-structured data is a real-world graph, namely the Texas network, where color indicates the labels of nodes.
We obtain a re-connected graph through attention score computing, sorting, and screening.
Subsequently, the re-connected graph is optimized together with the evolution of node embeddings.
Finally, we obtain a re-connected graph with high homophily}.
\label{similar}  
\end{figure*}

In this study, the re-connected graph and node embeddings are optimized simultaneously and benefit each other.
In this manner, we can obtain a sparse, nonnegative, connected, and reliable re-connected adjacency matrix ${\bf A}_{*}$. 
Figure~\ref{similar} shows the construction process of the re-connected adjacency matrix.
In summary, the MHSA mechanism allows nodes to interact with any other node, and appropriate screening mechanisms help the nodes identify the most relevant new neighbors.

\subsection{Modified Transformer Block}\label{AM}
Transformer blocks with self-attention mechanisms can help the model recognize important features from the entire graph and pay more attention to these features~\cite{self-atten}.
Thus, the transformer block is an appropriate method to help the model capture long-range dependencies.
However, using a transformer block directly to help GCN perform feature fusion can result in over-fitting.
Therefore, we modify the transformer block to make it more applicable to GCN. A modified transformer block is shown in Figure~\ref{fig:Fig.1}.
The MHSA mechanism of the modified transformer block is represented as:
\begin{equation}
{\bf H}_{\text {MHSA}}=\text{Concat}\left(\text {head}_{1},\text{head}_{2},\ldots,\text{head}_{\mathrm{m}}\right),
\label{concat}
\end{equation}
\begin{equation}
\text { head }_{i}=\operatorname{softmax}\left(\frac{{\bf Q}_{i} {\bf K}_{i}^{\mathrm{T}}}{\sqrt{d_{k}}}\right) {\bf V}_{i},
\label{innerproduct}
\end{equation}
\begin{equation}
{\bf Q}_{i}={\bf H}_{\text {original}}{\bf W}_{i}^{Q},
\label{Q}
\end{equation}
\begin{equation}
{\bf K}_{i}={\bf H}_{\text {original}}{\bf W}_{i}^{K},
\label{Ki}
\end{equation}
\begin{equation}
{\bf V}_{i}={\bf H}_{\text {original}}{\bf W}_{i}^{V},
\label{V}
\end{equation}
\begin{equation}
{\bf H}_{\text {original}}=\text{ReLU}({\bf X}{\bf W}^{0}+{\bf b}^{0}),
\label{Horiginal}
\end{equation}
where ${\bf W}^{0}\in \mathbb{R}^{d \times q}$, ${\bf W}_{i}^{Q}$, ${\bf W}_{i}^{K}$, and ${\bf W}_{i}^{V}\in \mathbb{R}^{q \times q/m}$ are learnable weighted matrix.
${\bf Q}_{i}$, ${\bf K}_{i}$ and ${\bf V}_{i}$ are the $Q$ (Query), $K$ (Key) and $V$ (Value) matrices derived from the linear transformation of ${\bf H}_\text{original}$, respectively.
Compared with ${\bf X}_\text{MHSA}$ in Eq.~\ref{concat0}, ${\bf H}_{\text {MHSA}}$ in Eq.~\ref{concat} omit a fully connected layer for reducing model complexity and minimizing information loss.
Then, We apply a dropout to ${\bf H}_\text{MHSA}\in \mathbb{R}^{n \times q}$ and ${\bf H}_\text{original}\in \mathbb{R}^{n \times q}$.

In the original transformer block~\cite{self-atten}, the MHSA mechanism is followed by a residual connection.
The residual connection can overcome the problem of gradient disappearance, enabling the development of deeper models.
Thus, we also apply the residual connection to ${\bf H}_{\text {MHSA}}$.
The output of the modified transformer block, fusional feature vectors, is calculated as follows:
\begin{equation}
{\bf H}_{\text {fusion}}= \text {Dropout}({\bf H}_{\text {original}}+{\bf H}_{\text {MHSA}})+{\bf H}_{\text {original}},
\label{Hnovel}
\end{equation}
where the purpose of using dropout is to alleviate the over-fitting problem.
For convenience, we represent the output of the modified transformer block as follows:
\begin{equation}
{\bf H}_{\text {fusion}}= \text{Modified\_Transformer\_Block}({\bf X}).
\label{TFF}
\end{equation}

\subsection{GCN-SA}\label{gcn-ef}

The fusional feature vectors and original feature vectors are concatenated as the ego-embeddings for nodes:
\begin{equation}
{\bf H} = {\bf H}_\text{fusion}\|{\bf H}_\text{original},
\label{ego-embedding}
\end{equation}
where $\|$ represents the concatenation function.
Afterward, we employ the ${\bf A}_{*}$ generated from Subsection~\ref{adjacent} to perform feature aggregation as follows:
\begin{equation}
{\bf H}_{A_{*}} = {\hat{\bf A}_{*}}{\bf H},
\label{eq:2-GCN-EF1}
\end{equation}
where ${\hat{\bf A}_{*}} ={\bf D}_{*}^{-1/2}{\bf A_{*}}{\bf D}_{*}^{-1/2}$ is the normalized ${\bf A_{*}}$, and ${\bf D}_* = \operatorname{diag}\left( {\bf A}_* {\bf 1}_n\right)$ is the degree matrix of ${\bf A}_*$.
We call the results of feature aggregation, ${\bf H}_{A_{*}}$, the reconnected-neighbor-embeddings.
Similarly, we utilize the original adjacency matrix ${\bf A}$ to perform feature aggregation as follows:
\begin{equation}
{\bf H}^{(k)}_{A} = \hat{{\bf A}}{\bf H}^{(k-1)}_{A},
\label{eq:2-GCN-EF}
\end{equation}
where $k=1,2,\ldots,K$, $K$ represents the times of the feature aggregation.
${\bf H}^{(0)}_{A} = {\bf H}$.
$\hat{{\bf A}} = {\bf D}^{-1/2}\left( {\bf A}+{\bf I}\right){\bf D}^{-1/2}$ represents the normalized ${\bf A}$ with self-loop.
Herein, ${\bf D} = \operatorname{diag}\left( \left({\bf A+I}\right) {\bf 1}_n\right)$ is the degree matrix of ${\bf A}$ with self-loop, and ${\bf I}\in \mathbb{R}^{n \times n}$ is a identity matrix.
${\bf H}^{(K-1)}_{A}$ and ${\bf H}^{(K)}_{A}$ are defined as the neighbor-embeddings of nodes.
Subsequently, the ego-embeddings, neighbor-embeddings, and re-connected neighbor-embeddings are concatenated as the general embeddings of nodes. The formula is written as:
\begin{equation}
{\bf H}^{cb} = \operatorname{Concat}({\bf H},
{\bf H}^{(K-1)}_{A},{\bf H}^{(K)}_{A}, {\bf H}_{A_{*}}).
\label{eq:2-GCN-EF2}
\end{equation}
Then we apply a dropout to ${\bf H}^{cb}$.
For graphs under homophily, ${\bf H}^{(K-1)}_{A}$ and ${\bf H}^{(K)}_{A}$ are sufficient for downstream task. This can be proved by GCN~\cite{{GCN+chapter2017}} and GAT~\cite{{GAT+chapter2018}}.
In addition, ${\bf H}_{A_{*}}$ can be treated as the supplement to ${\bf H}^{(K-1)}_{A}$ and ${\bf H}^{(K)}_{A}$.
For the graphs with low/medium homophily, ${\bf H}$ and ${\bf H}_{A_{*}}$ can also perform well.
Moreover, the approach of neighbor-embedding updating allows GCN-SA to capture higher-order neighbor information flexibly.

Afterward, a learnable weight vector ${\bf w}$ is generated, and the dimension of ${\bf w}$ is the same as the output dimensions of ${\bf H}^{cb}$.
Then we take the Hadamard product between ${\bf H}^{cb}$ and ${\bf w}$ to highlight the important dimension of ${\bf H}^{cb}$:
\begin{equation}
{\bf H}^{(1)} = \text{ReLU}({\bf w} \odot {\bf H}^{cb}).
\label{hadamard}
\end{equation}
Subsequently, nodes are classified in the following way:
\begin{equation}
{\bf Z}=\displaystyle softmax\left(\text{Modified\_Transformer\_Block}({\bf H}^{(1)}{\bf W}^{1})\right),
\label{TRANSFORMER2}
\end{equation}
where ${\bf W}^{1}$ is a trainable weighted matrix, and the output dimension of ${\bf W}^{1}$ is the same as the number of classes. The modified transformer block performing node embedding fusion can help GCN-SA fuse valuable information from different nodes to improve classification results. Then, we calculate the cross-entropy error over all labeled nodes:
\begin{equation}
\mathcal{L}_{\text {pred}}=-\sum_{i \in \mathcal{Y}_{lab}} \sum_{j=1}^{c} Y_{i j} \ln Z_{i j},
\label{eq:2-GCN-EF4}
\end{equation}
where $\mathcal{Y}_{lab}$ is the set of node indices that have labels.

\begin{algorithm}[tb]
   \caption{GCN-SA}
   \label{alg:example}
\begin{algorithmic}
   \STATE {\bfseries Input:} ${\bf X}$: feature matrix of nodes, ${\bf Y}_{lab}$: label matrix of nodes with labels,
   ${\bf A}$: original adjacency matrix, ${\bf w}$: trainable weighted vector,
   ${\bf W}_{i}^{K}$, ${\bf W}_{i}^{Q}$, ${\bf W}_{i}^{V}$, ${\bf W}^{0}$, ${\bf W}^{1}$ : trainable weighted matrixs.
   \STATE {\bfseries Output:} ${\bf Z}$: probability distribution of nodes.
   \STATE 1. ${\bf A}_{*}\leftarrow \text{structure learning} ({\bf X})$ in~\ref{Re-connected}.
   \STATE 2. ${\bf H}_\text{original}\leftarrow \left \{ {\bf X}, {\bf W}^{0}\right \}$ in Eq.~\ref{Horiginal}
   \STATE 3. ${\bf H}_\text{fusion}\leftarrow \left \{{\bf H}_\text{original}, {\bf W}_{i}^{K}, {\bf W}_{i}^{Q}, {\bf W}_{i}^{V}\right \}$in~\ref{AM}
   \STATE 4. ${\bf H} = \text{Concat}({\bf H}_\text{original},{\bf H}_\text{fusion})$
   \STATE 5. ${\bf H}_{A_{*}}\leftarrow \left \{ {\bf H}, {\bf A}_{*}\right \}$ in Eq.~\ref{eq:2-GCN-EF1}.
   \STATE {\bfseries repeat} (initialize $k$ to 1)
   \STATE 6. Update ${\bf H}^{(k)}_{A}\leftarrow \left \{ {\bf H}^{(k-1)}_{A},{\bf A}\right \}$, ${\bf H}^{(0)}_{A}={\bf H}$ in Eq.~\ref{eq:2-GCN-EF},\\ ~~~$k=k+1$.
   \STATE {\bfseries until $k=K$}
   \STATE 7. ${\bf H}^{cb} \leftarrow \left \{{\bf H},{\bf H}^{(K-1)}_{A},{\bf H}^{(K)}_{A}, {\bf H}_{A_{*}}\right \}$ in Eq.~\ref{eq:2-GCN-EF2}.
   \STATE 8. ${\bf H}^{(1)} \leftarrow \left \{{\bf H}^{cb}, {\bf w}\right \}$ in Eq.~\ref{hadamard}.
   \STATE 9. ${\bf Z} \leftarrow \left \{{\bf H}^{(1)}, {\bf W}^{1} \right \}$ in Eq.~\ref{TRANSFORMER2}.
   \STATE 10. $\mathcal{L}_{pred}\leftarrow $ LOSS $({\bf Z}_{lab}, \mathcal{Y}_{lab})$ in Eq.~\ref{eq:2-GCN-EF4}.
   \STATE 11. Backpropagate $\mathcal{L}_{pred}$ to update model weights.
\end{algorithmic}
\end{algorithm}

We employ an MHSA mechanism to capture the internal correlation between nodes and construct a fully connected re-connected adjacency matrix.
Afterward, We obtain a connected and sparse re-connected adjacency matrix by masking the unnecessary edges.
Subsequently, we obtain a reliable, connected, and sparse re-connected adjacency matrix through joint learning of the re-connected graph and node embeddings.
Moreover, we propose a modified transformer block and use it to perform feature vector fusion.
Next, we combine the fusional feature vector with the node feature vectors to train the model and reuse the modified transformer block to perform embedding fusion to help GCN-SA select valuable features from different nodes.
Our GCN-SA model is trained using diverse information from different views to learn representative node embeddings and accurately perform downstream tasks.
These steps can help GCN-SA capture long-range dependencies, allowing it to adapt to graphs with varying levels of homophily.
The pseudo-code for the proposed GCN-SA is provided in Algorithm~\ref{alg:example}.

\section{Experimental Results}
In the experimental section, to demonstrate the merits of our GCN-SA, we compare it with several outstanding GNNs on node classification tasks.
The experiments are conducted on eight benchmark datasets with different homophily levels.
Subsection~\ref{Datasets} describes the graph datasets considered in this study.
Subsection~\ref{Baselines} discusses the eight baseline techniques used in this study.
Subsection~\ref{Experimental Setup} illustrates the experimental setup.
Subsection~\ref{evram} evaluates the effectiveness of the reconnected adjacency matrix.
Subsection~\ref{emtba} investigates the impact of the modified transformer block on the model.
Subsection~\ref{ Complexity} analyzes the time complexity of our GCN-SA.
Subsection~\ref{cadg} estimates the classification performance of our 
GCN-SA and other competitors in the literature.

\subsection{Datasets}\label{Datasets}
Eight open graph datasets are adopted to validate the proposed GCN-SA in the simulation, including three citation networks, two Wikipedia networks, and three WebKB networks.

The three standard citation network benchmark datasets include the Cora, Citeseer, and Pubmed datasets~\cite{{PR2}}.
In the three citation networks, nodes correspond to papers, and node features are defined as the bag-of-words representation of the paper. Edges correspond to citations between papers. Each node has a unique label, defined as the corresponding paper's academic topic.

Wikipedia networks are page-page networks on specific topics in Wikipedia, e.g., Chameleon and Squirrel datasets.
In Wikipedia networks, nodes correspond to Wikipedia pages, and node features are represented as informative nouns on Wikipedia pages.
Edges correspond to the reciprocal links between Wikipedia pages.
Nodes are classified into five categories according to the amount of their average traffic.

The sub-networks of the WebKB networks include the Cornell, Texas, and Wisconsin datasets.
These three datasets were collected from the computer science departments~\cite{{Geom+chapter2020}}.
In WebKB networks, nodes correspond to web pages, and the bag-of-words representation of web pages serves as the node features.
Edges represent hyperlinks between web pages. These web pages are divided into five classes.
\begin{table*}[t]\small
\caption{Summary of the datasets utilized in our experiments.}
\label{Tab:des-datasets}
\begin{center}
\vspace{-2mm}
\setlength\tabcolsep{3pt}
\begin{tabular}{l c c c c c c c c }
\toprule
Dataset& Cora & Citese. & Pubmed & Chamele. &  Squirr.& Cornell & Texas & Wiscon. \\ \hline  
\ Hom.ratio $h$   & 0.81 & 0.74  & 0.8   &0.23   &0.22 &0.3 &0.11&0.21\\
\# Nodes           & 2708 & 3327 & 19717 &2277   &5201 &183 &183 &251\\
\# Edges           & 5429 & 4732 & 44338 &31421  &198493&295 &309 &499\\
\# Features        & 1433 & 3703 & 500   &2325   &2089 &1703&1703&1703\\
\# Classes         & 7    & 6    & 3     &5      &5    & 5  &5   &5  \\
\bottomrule
\end{tabular}
\end{center}
\end{table*}

Unless particularly specified, nodes per class are randomly split into $60\%$, $20\%$, and $20\%$ for training, validation, and testing by default across all datasets.
Testing is performed when validation losses reach a minimum.
A summary of the information for all datasets is presented in Table~\ref{Tab:des-datasets}.

The homophily of graph-structured data is a crucial characteristic and plays a significant role in analyzing and utilizing them.
As in~\cite{hhgcn}, we also utilize the homophily ratio to measure the homophily level of a graph,
$h=\frac{\left|\left\{(u, v):(u, v) \in \mathcal{E} \wedge y_{u}=y_{v}\right\}\right|}{|\mathcal{E}|}$, where $\mathcal{E}$ represents the set of edges,
$y_{u}$ and $y_{v}$ represent the labels of node $u$ and $v$, 
respectively.
$h$ is the proportion of edges in a graph that connect nodes with the same label (i.e., intra-class edges).
This definition is proposed in~\cite{hhgcn}.
Graphs have strong homophily if $h$ is high ($h \rightarrow 1$) and weak homophily if $h$ is low ($h \rightarrow 0$).
Table~\ref{Tab:des-datasets} lists the homophily ratio $h$ for each graph.
From the $h$ values of the adopted graphs, we can observe that all citation networks display a high level of homophily, while the WebKB networks and Wikipedia networks exhibit low homophily.

\subsection{Baselines}\label{Baselines}
We compare the proposed GCN-SA with several baselines:

$\bullet$ MLP~\cite{MLP} is a simple deep neural network. MLP makes predictions based on the features of nodes without considering any local or non-local information.

$\bullet$ GAT~\cite{GAT+chapter2018} enables specifying different weights to different neighbors by employing the attention mechanism.

$\bullet$ GCN~\cite{GCN+chapter2017} is one of the most common GNNs.
The GCN is introduced in Subsection~\ref{GCN}. GCN makes predictions by aggregating local information.

$\bullet$ MixHop~\cite{MixHop} repeatedly aggregates the feature representations of neighbors with various distances to learn robust node
embeddings.

$\bullet$ Geom-GCN~\cite{Geom+chapter2020} captures geometric information of graphs to enhance GCN's ability on representation learning.
Three embedding methods, struc2vec, Isomap, and Poincare embedding, are employed in Geom-GCN, corresponding to three variants: Geom-GCN-S, Geom-GCN-I, and Geom-GCN-P. Nevertheless, we only report the best results of three Geom-GCN variants.

$\bullet$ IDGL~\cite{IDGL} jointly and iteratively learning the node embeddings and the graph structure.
In addition, IDGL dynamically stops when the learned graph is close enough to the optimized graph for the downstream prediction task.

$\bullet$ H$_{2}$GCN~\cite{hhgcn} identifies a set of significant designs, including ego and neighbor embedding separation, higher-order neighborhoods.
By colligating these designs, H$_{2}$GCN adapts to both heterophily and homophily.
We consider two variants: H$_{2}$GCN-1 and H$_{2}$GCN-2 according to the distance of neighbor nodes for each aggregation.

$\bullet$ CPGNN~\cite{CPGNN} incorporates an interpretable compatibility matrix for modeling the homophily level in the graph, which enables it to go beyond the assumption of strong homophily.
CPGNN contains four variants: CPGNN-MLP-1, CPGNN-MLP-2, CPGNN-Cheby-1 and CPGNN-Cheby-2.
According to the analysis in \cite{CPGNN}, CPGNN-Cheby-1 performs the best overall.
Therefore, we consider the CPGNN-Cheby-1.

\subsection{Experimental Setup}\label{Experimental Setup}
For comparison, we use eight state-of-the-art node classification algorithms, including
MLP, GAT, GCN, MixHop, GEOM-GCN, IDGL, H$_{2}$GCN and CPGNN.
In the above eight network models, 
all hyper-parameters are set according to \cite{hhgcn}, IDGL to~\cite{IDGL}, and CPGNN to \cite{CPGNN}.
Our GCN-SA includes many hyper-parameters, some of which are fixed, while others need to be searched on the validation set.
The random seed is set as $42$ for all experiments.
The number of attention heads $m$ is $4$ in the re-connected graph learning, and $m$ is $1$ in feature vector and embedding fusion.
The numbers of hidden features $p$ are set as $16$.
The descriptions of the hyperparameters need to be searched, and the corresponding ranges of values tried are provided in Table~\ref{hyperparameters}.
Table~\ref{hyperparameters_value} summarizes the values of all hyper-parameters and the corresponding classification accuracies on different datasets.

\begin{figure*}[t]
\centering
\subfigure{ \label{legend}
\includegraphics[width= 0.55in,height=1.05 in]{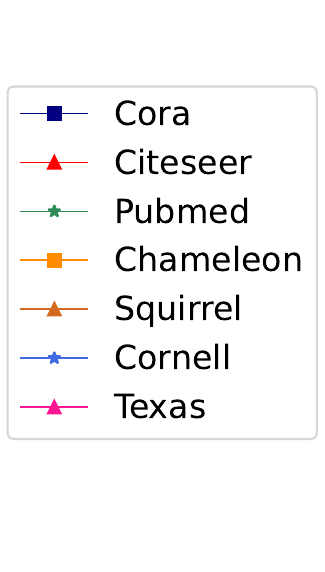}}
\hspace{-0.2cm}
\subfigure{\label{K} 
\includegraphics[width=1.33in,height=1.23in]{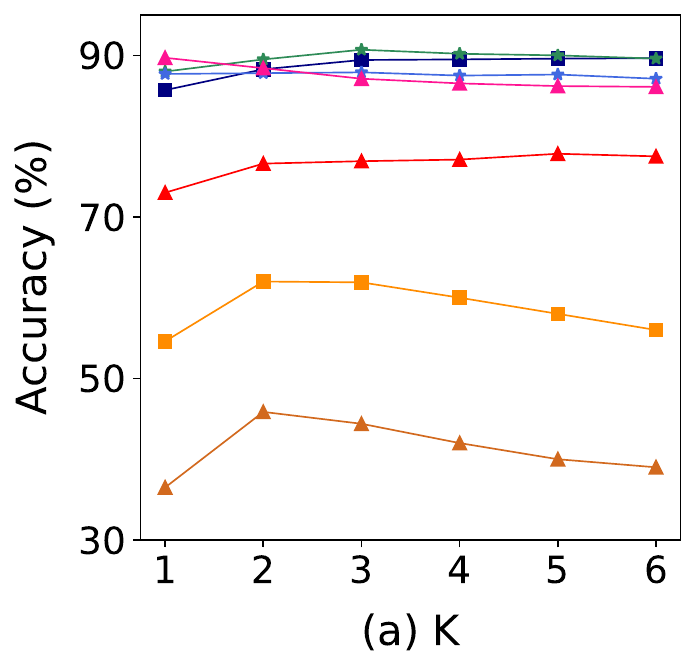}}
\hspace{-0.2cm}
\subfigure{ \label{r} 
\includegraphics[width=1.33in,height=1.23in]{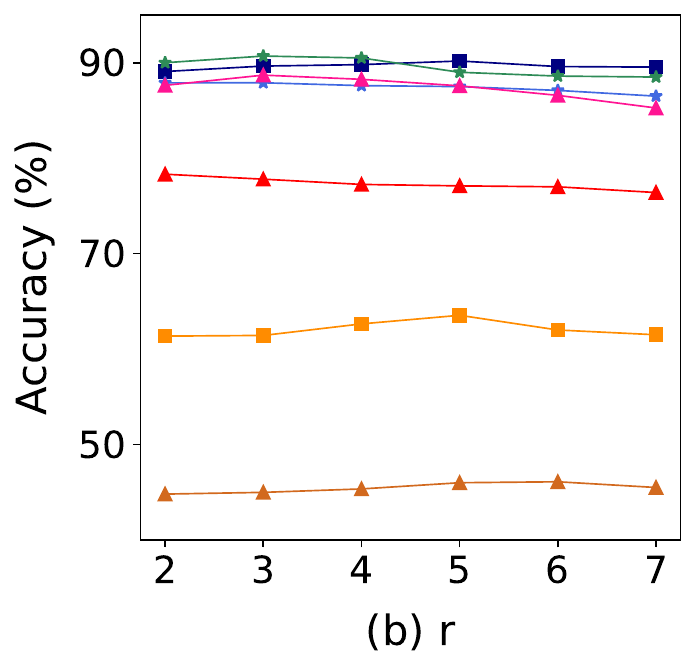}}
\hspace{-0.2cm}
\subfigure{ \label{e} 
\includegraphics[width=1.33in,height=1.23in]{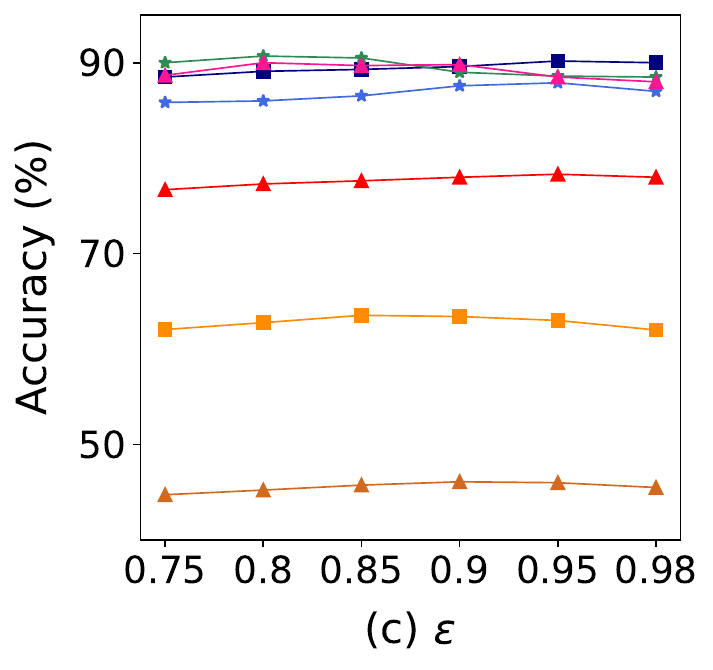}}
\vspace{-0.3cm}
\caption{Node classification accuracies (\%) of the proposed GCN-SA versus the hyper-parameters K, r and $\epsilon$ on seven datasets.}\vspace{-2mm}
\label{survey}  
\end{figure*}

Among all the hyper-parameters, $K$, $r$, and $\epsilon$ play a particularly crucial role in our GCN-SA.
$K$ determines the number of times feature aggregation is performed on the original graph structure, while $r$ and $\epsilon$ dictate the sparsity and connectivity of the reconnected graph.
Therefore, we investigate the impact of these three hyper-parameters on the classification accuracies of our GCN-SA on seven datasets.
Figure~\ref{survey} shows the classification accuracies versus the $K$, $r$, and $\epsilon$ on different datasets.
Since the behavior of GCN-SA in the Texas and Wisconsin datasets is similar, we omit the Wisconsin dataset for brevity.
Referring to Figure~\ref{survey} (b), it can be observed that as $K$ increases, GCN-SA behaves differently across various graphs. 
This phenomenon is mainly because different datasets have different levels of homophily.
As for Figure~\ref{survey} (c) and (d), the classification accuracies of our GCN-SA remain relatively stable with increasing values of $r$ and $\epsilon$. This consistency indicates that our graph-structure-learning method is effective and robust. The optimal values of $K$, $r$, and $\epsilon$ for different datasets are recorded in the Table~\ref{hyperparameters_value}.

ReLU is utilized as the nonlinear activation function.
The training nodes are utilized for training our GCN-SA by minimizing cross-entropy loss.
Adam optimizer~\cite{adam} is employed in our GCN-SA.
The proposed GCN-SA is implemented with the deep learning library PyTorch.
The Python and PyTorch versions are 3.8.10 and 1.9.0, respectively.
All experiments are conducted on a Linux server with a GPU (NVIDIA GeForce RTX 2080 Ti).
The experimental results are the mean and standard deviation of ten runs for all the experiments.

\subsection{Effectiveness Verification for Re-connected Adjacency Matrix}\label{evram}

\begin{figure*}[t]
\centering 
\includegraphics[width=4.8in,height=1.58in]{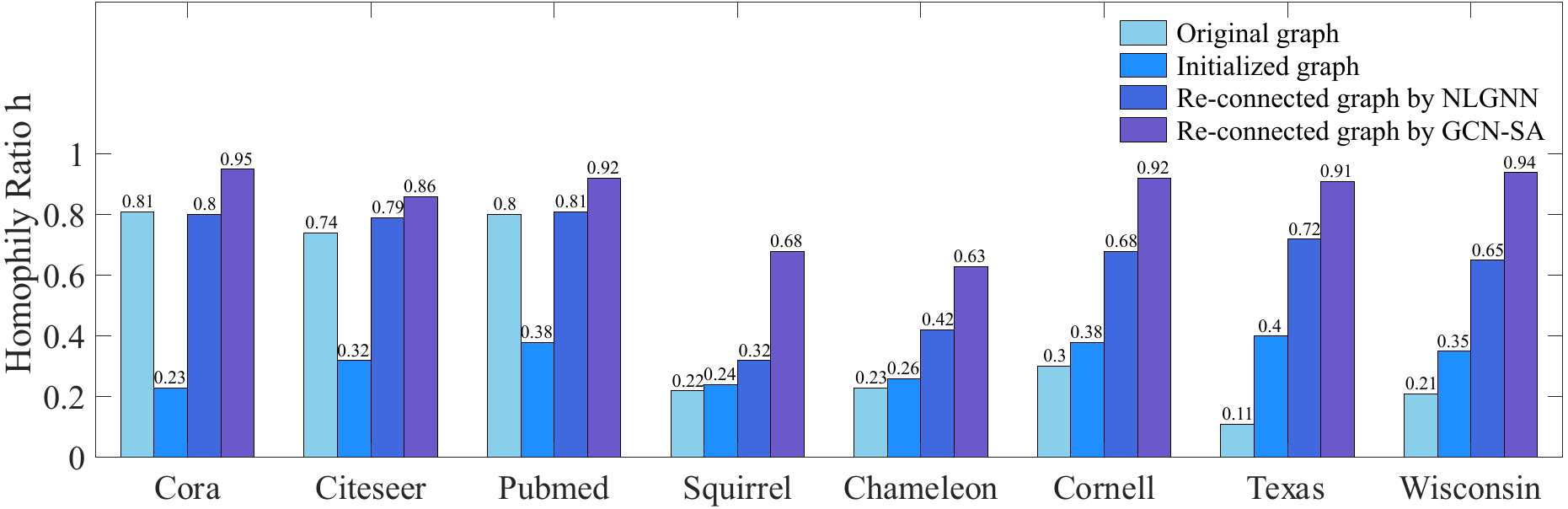}
\vspace{-0.3cm}
\caption{Comparisons of homophily ratio $h$ for graphs, including the original, initialized, and re-connected graphs.}
\label{homo}  
\end{figure*}

\begin{figure*}[t]
\centering
\subfigure[\hspace{-0.4cm}][Original graph]{ \label{oldgraph}
\includegraphics[width=1.65in,height=1.65in]{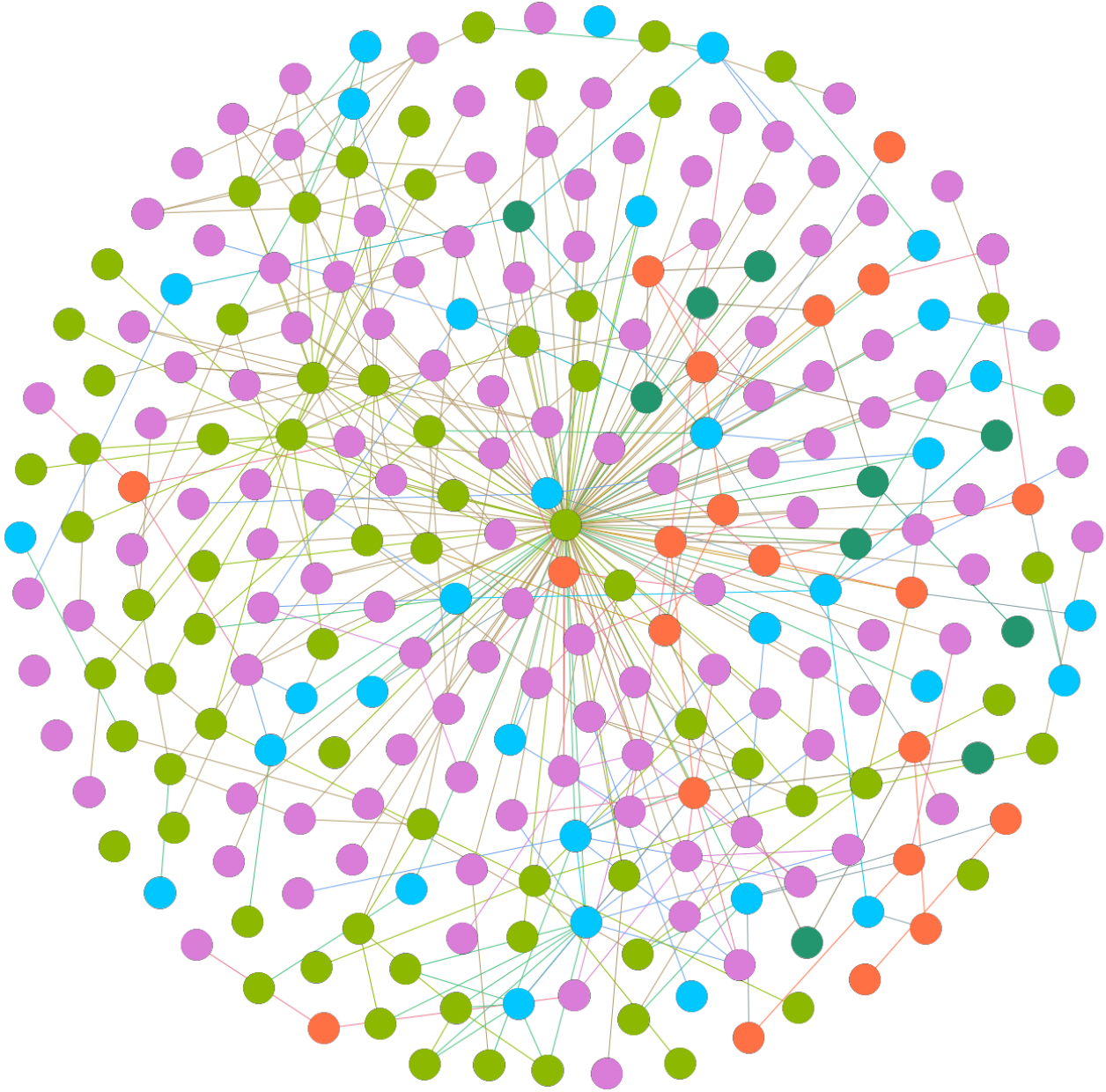}}
\hspace{-0.2cm}
\subfigure[\hspace{-0.4cm}][Neighbors in original graph]{ \label{neighborsinoldgraph} 
\includegraphics[width=1.65in,height=1.65in]{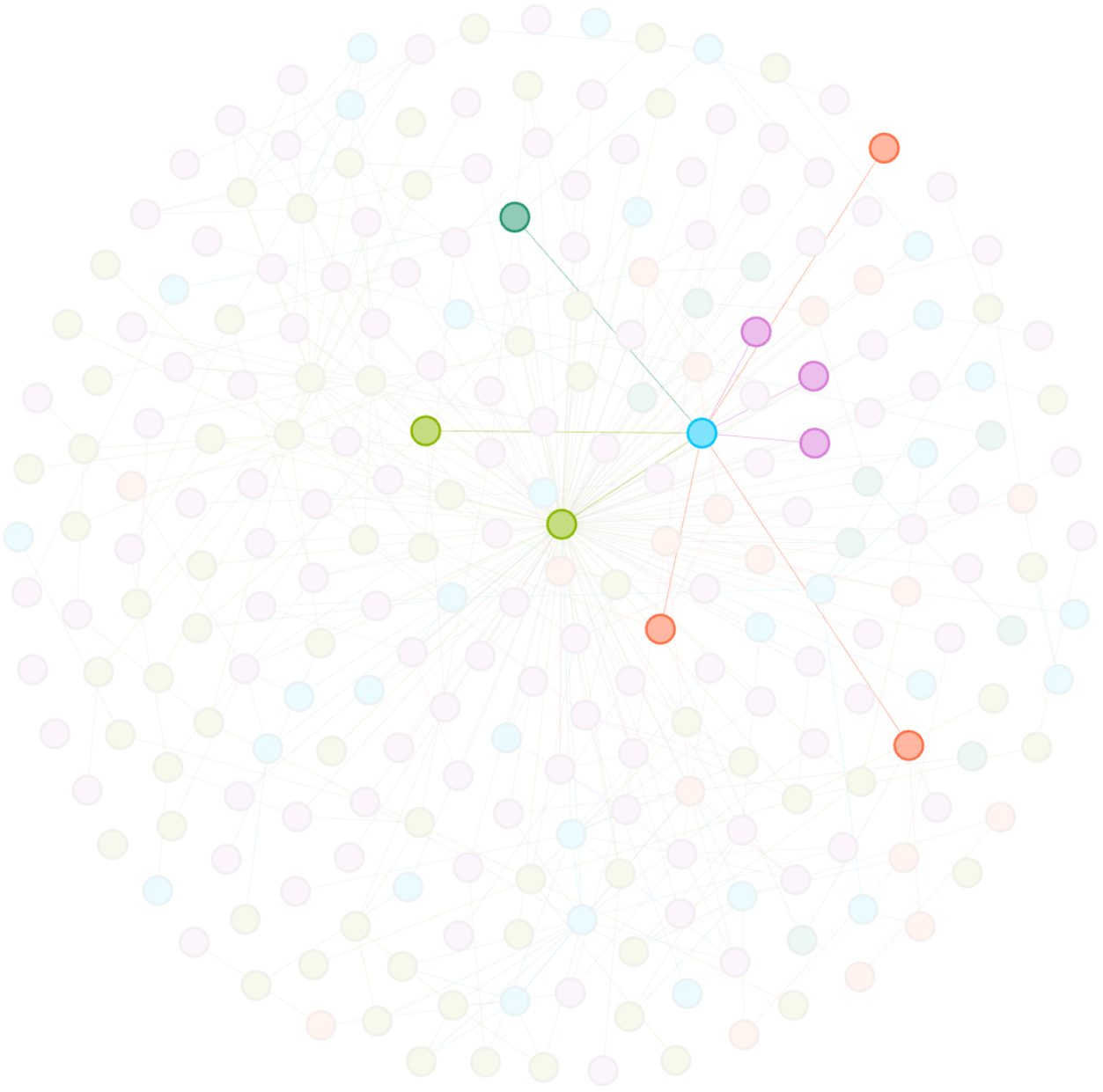}}
\hspace{-0.2cm}
\subfigure[\hspace{-0.4cm}][Re-connected graph]{ \label{newgraph} 
\includegraphics[width=1.65in,height=1.65in]{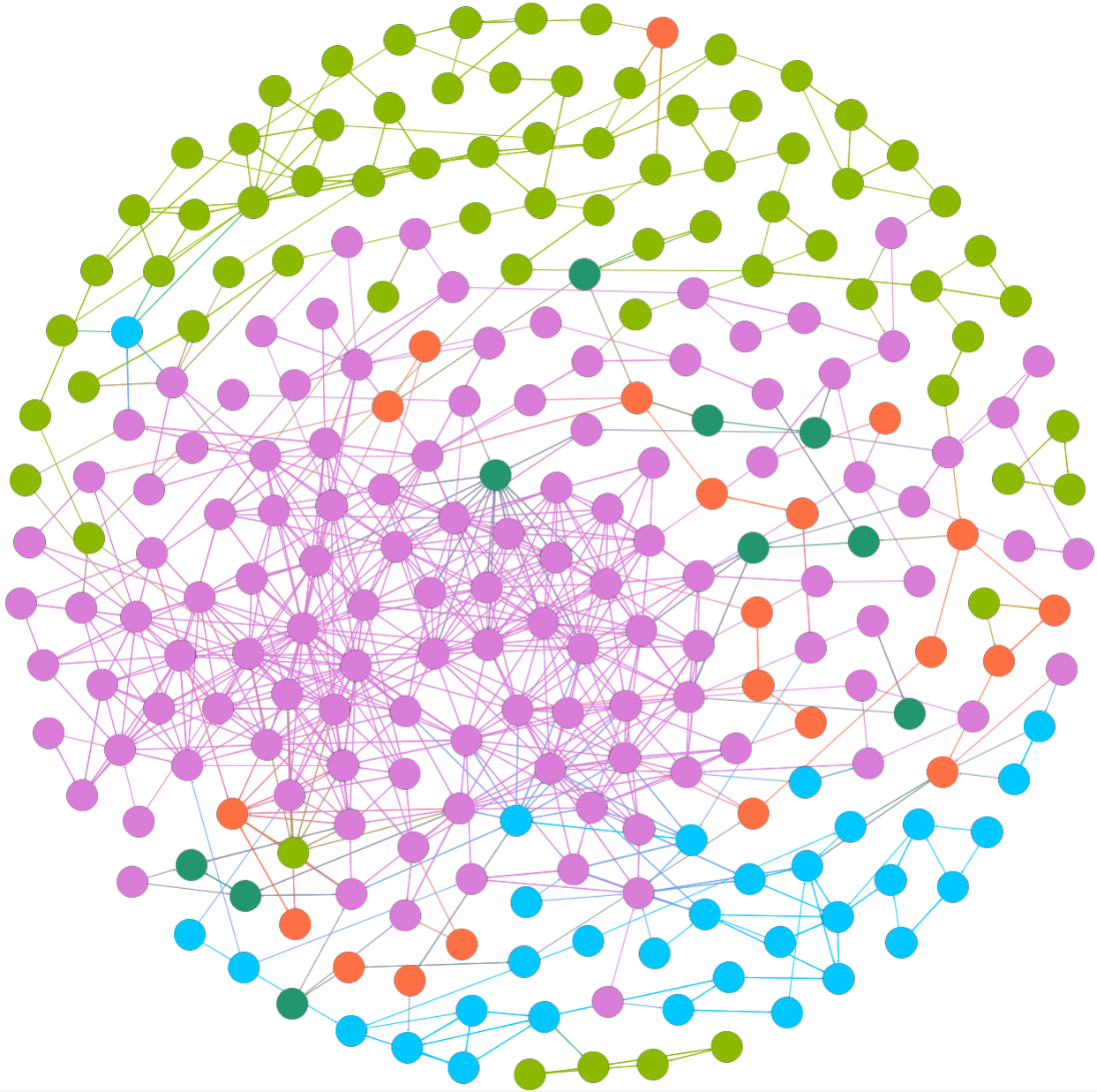}}
\hspace{-0.2cm}
\subfigure[\hspace{-0.4cm}][Neighbors in re-connected graph]{ \label{neighborsinnewgraph} 
\includegraphics[width=1.65in,height=1.65in]{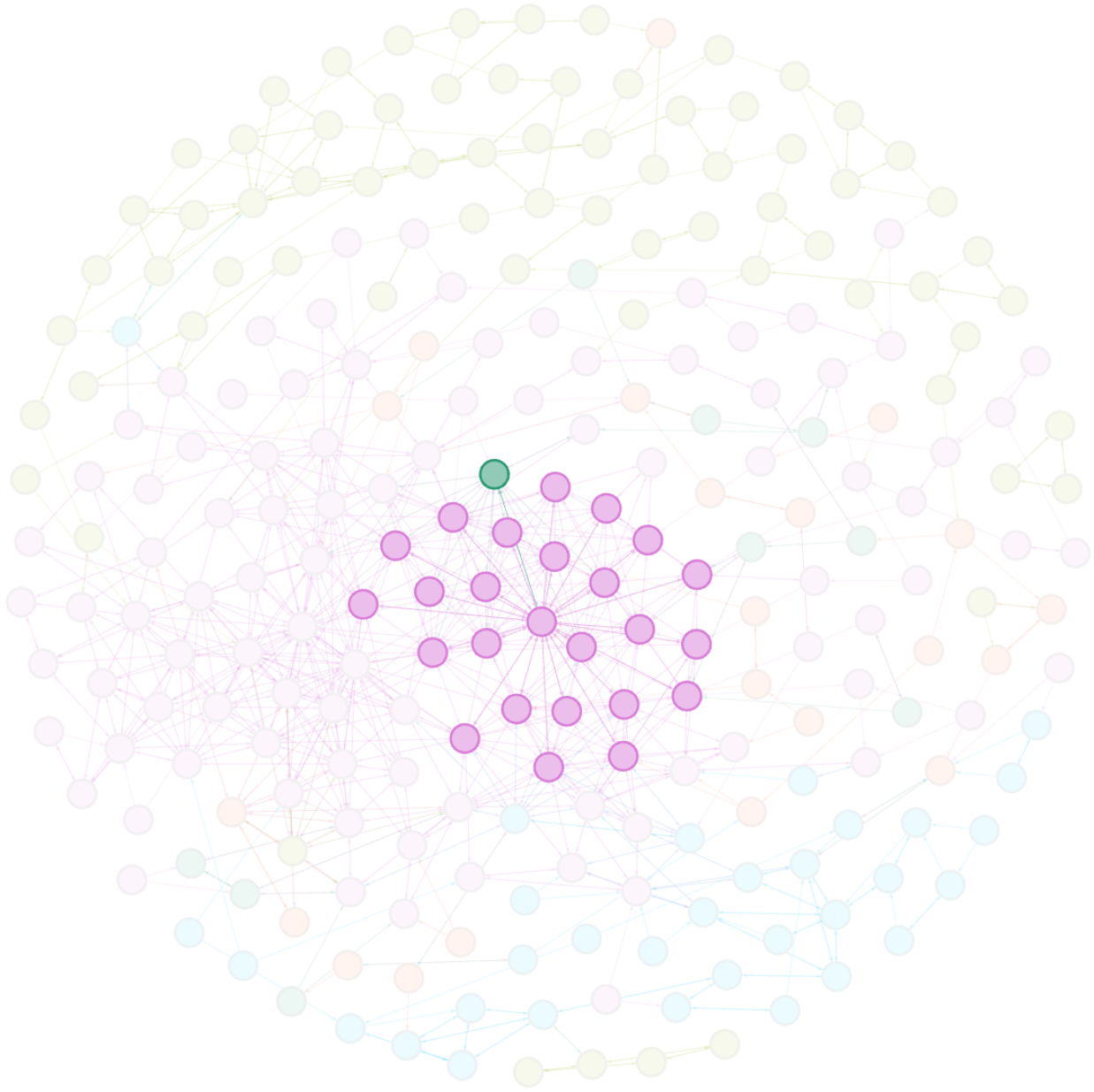}}
\vspace{-0.3cm}
\caption{Visualization of the graph structures. The graph-structured data is a real-world graph, namely the Wisconsin network, where color indicates the labels of nodes. The original graph is shown in (a), and the re-connected graph learned by GCN-SA is shown in (c).
We fade the remaining nodes and edges to emphasize the neighbors of the selected node in (b) and (d).}\vspace{-2mm}
\label{gephi}  
\end{figure*}

In this section, we investigate the impact of our GCN-SA on the reliability of re-connected graph structures. 
We employ the homophily ratios, denoted as $h$, to evaluate the reliability of the reconnected graph structures.
Higher values of $h$ correspond to greater reliability, while lower values indicate less reliability.
Figure~\ref{homo} displays the $h$ values for the original, initialized, and two reconnected graphs across eight datasets. 
The initialized graph is constructed by GCN-SA without optimization.
In contrast, the two reconnected graphs are learned and optimized by NLGNN and GCN-SA methods, respectively.
As seen from Figure~\ref{homo}, 
the $h$ values of the re-connected graphs are significantly higher than those of the original and initialized graphs for all datasets, particularly in the case of Wikipedia and WebKB networks. 
This is mainly because the re-connected graphs can be optimized toward the downstream prediction task in both NLGNN and GCN-SA methods.
Additionally, we observe that the $h$ values of the re-connected graphs learned by our GCN-SA are always higher than those of the NLGNN.
This difference is due to the limited power of the inner product-based similarity metric function employed by NLGNN in representing the intrinsic correlation between nodes. In comparison, our GCN-SA is more effective at capturing the internal correlation of nodes.

\begin{table*}[t]
\caption{Node classification accuracies (\%) of the proposed GCN-SA.}
\vspace{-0.4cm}
\label{re-connected}
\begin{center}
\scriptsize
\setlength\tabcolsep{1.5pt}
\begin{tabular}{l c c c c c c c c c c c }
\toprule
Dataset&${\bf A}$& ${\bf A}_{*}$ &Cora & Cite. & Pubm. &  Cham. &Squi. & Corn. & Texa. & Wisc.\\ \hline
              &--&--&75.7$\pm$1.8 &72.8$\pm$2.1 &86.8$\pm$0.8&49.8$\pm$3.2&34.3$\pm$2.0&84.7$\pm$6.3&85.0$\pm$5.5&87.4$\pm$4.5\\
GCN-SA&\checkmark&--&89.4$\pm$0.5 &77.0$\pm$0.8 &88.5$\pm$1.2&61.1$\pm$3.0&44.1$\pm$2.5&85.8$\pm$4.0&86.1$\pm$3.8&87.2$\pm$4.5\\
 &\checkmark&\checkmark&\bf{90.5$\pm$0.4}&\bf{79.1$\pm$0.6}&\bf{90.9$\pm$0.6}&\bf{64.1$\pm$1.8}&\bf{46.5$\pm$1.6}&\bf{90.8$\pm$4.5}&\bf{89.7$\pm$2.6}&\bf{91.5$\pm$2.5}\\
\bottomrule
\end{tabular}
\end{center}
\end{table*}

We employ the Gephi tool to visualize the original graph and the re-connected graph optimized by our GCN-SA, using the Wisconsin network as an example. 
Figure~\ref{gephi} presents the visualization of these two graphs,
allowing for an intuitive examination of the structural changes brought about by our GCN-SA. 
We then select a specific node and highlight its neighborhood in both the original graph (Figure~\ref{neighborsinoldgraph}) and the re-connected graph (Figure~\ref{neighborsinnewgraph}).
As depicted in Figure~\ref{gephi}, the original graph appears chaotic, with numerous inter-class connections.
However, after applying our GCN-SA, the structure of the reconnected graph becomes considerably more organized, with the fraction $h$ of intra-class edges reaching 90\%. 
This improvement in accuracy performance demonstrates that our graph-structure-learning module can substantially enhance the graph structure.

\begin{table*} 
\caption{Node classification accuracies (\%) of the proposed GCN-SA.}
\vspace{-0.3cm}
\label{fusion}
\begin{center}
\scriptsize
\setlength\tabcolsep{0.5pt}
\begin{tabular}{l c c c c c c c c c c c }
\toprule
Dataset&Fusion\_1& Fusion\_2& Cora & Cite. & Pubm. &  Cham. &Squi. & Corn. & Texa. & Wisc. \\ \hline
\multirow{4}{*}{GCN-SA}&--&--&89.0$\pm$0.7 & 76.9$\pm$1.1 & 89.7$\pm$1.3 & 61.4$\pm$3.5 & 42.5$\pm$3.0 & 86.1$\pm$5.0 & 86.5$\pm$4.5 & 87.1$\pm$4.8 \\
&\checkmark&--&89.4$\pm$0.6 & 77.5$\pm$1.0 & 90.1$\pm$1.2 &62.6$\pm$3.0 & 43.6$\pm$2.8 & 87.4$\pm$6.2 & 86.1$\pm$5.3 &88.2$\pm$4.8  &\\
&--&\checkmark& 89.7$\pm$0.4& 77.2$\pm$1.0 & 90.2$\pm$1.1 & 61.5$\pm$3.5 &43.5$\pm$2.7 & 86.8$\pm$5.3 & 89.0$\pm$4.9 & 89.2$\pm$5  \\
&\checkmark&\checkmark&\bf{90.5$\pm$0.4}&\bf{79.1$\pm$0.6}&\bf{90.9$\pm$0.6}&\bf{64.1$\pm$1.8}&\bf{46.5$\pm$1.6}&\bf{90.8$\pm$4.5}&\bf{89.7$\pm$2.6}&\bf{91.5$\pm$2.5}\\
\bottomrule
\end{tabular}
\end{center}
\end{table*}

\begin{table*}[t]
\caption{Classification accuracies (\%) of the proposed GCN-SA and two GCN-SA-based variants.}
\vspace{-0.3cm}
\scriptsize
\label{MTF}
\begin{center}
\setlength\tabcolsep{1.7pt}
\begin{tabular}{l c c c c c c c c }
\toprule
Method & Cora & Cites. & Pubmed &  Chame. &Squir. & Corne. & Texas & Wiscon. \\ \hline  
GCN-SA-W/O & 89.0$\pm$0.7 & 76.9$\pm$1.1 & 89.7$\pm$1.3 & 61.4$\pm$3.5 & 42.5$\pm$3.0 & 86.1$\pm$5.0 & 86.5$\pm$4.5 & 87.1$\pm$4.8 \\
GCN-SA-TF& 87.4$\pm$1.1 & 75.4$\pm$1.2 &87.2$\pm$ 1.5 & 58.5$\pm$5.0 & 41.2$\pm$2.5 & 83.9$\pm$6.8 & 84.5$\pm$6.4&85.2$\pm$5.8\\
GCN-SA&\bf{90.5$\pm$0.4}&\bf{79.1$\pm$0.6}&\bf{90.9$\pm$0.6}&\bf{64.1$\pm$1.8}&\bf{46.5$\pm$1.6}&\bf{90.8$\pm$4.5}&\bf{89.7$\pm$2.6}&\bf{91.5$\pm$2.5}\\
\bottomrule
\end{tabular}
\end{center}
\end{table*}

Afterward, we investigate the impact of the ${\bf A}_{*}$ on the accuracy of our GCN-SA using an ablation study.
Table~\ref{re-connected} presents the accuracies of our GCN-SA across all adopted graphs.
It can be observed that $\bf{A}$ significantly enhances the classification accuracies of our GCN-SA in citation networks and Wikipedia networks.
This improvement is due to the $\bf{A}$, which can provide valuable neighbors for aggregation.
Additionally, $\bf{A}_{*}$ also promotes the performance of our GCN-SA, particularly in WebKB networks.
This is owing to $\bf{A}_{*}$ being more reliable than $\bf{A}$ in WebKB networks.
Our GCN-SA, when combined with both $\bf{A}$ and $\bf{A}{*}$, performs well in all networks.
Based on this observation, ${\bf A}_{*}$ can be treated as a qualified adjacency matrix for providing each node with reliable neighbors, thereby enhancing the ability of GCN-SA to capture long-range dependencies.

\subsection{Effect of the Modified Transformer Block on Accuracy}\label{emtba} 

In this section, we examine the impact of the modified transformer block on the accuracies of GCN-SA. 
The modified transformer block operates at two different points within GCN-SA: feature vector fusion (Fusion\_1) and node embedding fusion (Fusion\_2).
We employ an ablation study to assess the effects of Fusion\_1 and Fusion\_2 on the node classification accuracy of GCN-SA.
Table~\ref{fusion} displays the classification accuracies of GCN-SA across all datasets. As observed, Fusion\_1 enhances the performance of GCN-SA. For example, when Fusion\_1 is applied to GCN-SA in the Chameleon dataset, the performance improves by 1.2\%. Similarly, Fusion\_2 also contributes to the improvement of GCN-SA's accuracies. In the Texas dataset, GCN-SA achieves a 2.5\% performance enhancement when Fusion\_2 is incorporated.
Ultimately, GCN-SA with both Fusion\_1 and Fusion\_2 achieves the best performance across all networks. Notably, in the Squirrel dataset, GCN-SA sees a 3.6\% accuracy improvement when both Fusion\_1 and Fusion\_2 are applied. The experimental results demonstrate that both Fusion\_1 and Fusion\_2 can bolster GCN-SA's ability to learn node embeddings effectively.

For further insight, we compare the impacts of the original transformer block and modified transformer block on the accuracies of GCN-SA. 
For this purpose, we introduce two variants of GCN-SA.
The first variant, GCN-SA-W/O, that without feature vector and embedding fusion.
The second variant, GCN-SA-TF, employs the original transformer block to perform feature vector and node embedding fusion.
Table~\ref{MTF} displays the classification accuracies of GCN-SA-W/O, GCN-SA-TF, and GCN-SA across all datasets.
As observed, GCN-SA-TF performs the worst, even underperforming GCN-SA-W/O.
This might be due to the original transformer block making GCN suffer from over-fitting.
In contrast, GCN-SA achieves the best classification performance, demonstrating that the modified transformer block is more effective in enhancing GCN compared to the original transformer block.

\begin{figure*}[t]
\centering 
\includegraphics[width=4.8in,height=1.48in]{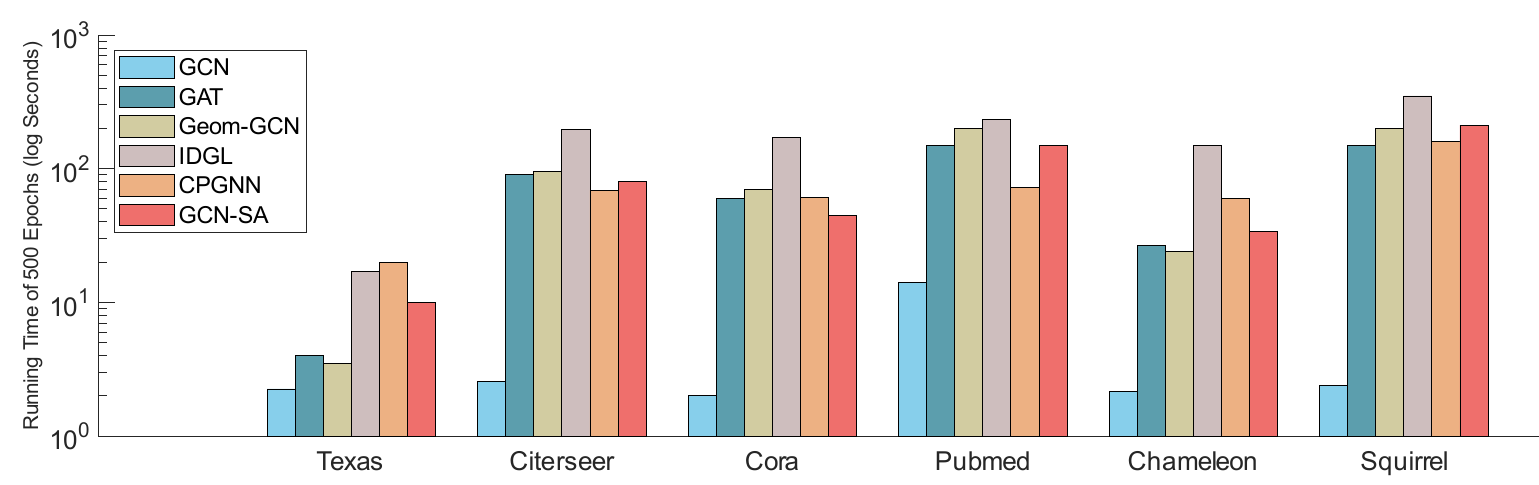}
\vspace{-3mm}
\caption{Running time comparison. GCN, GAT, Geom-GCN, IDGL, CPGNN, and GCN-SA run 500 epochs, and the y-axis is the log seconds. GCN is the fastest, and GAT and GCN-SA are on the same level.}
\label{time}  
\end{figure*}

\begin{table*}[t]
\caption{Classification accuracies (\%) of the MLP, GAT, GCN, MixHop, Geom-GCN, IDGL, H$_{2}$GCN, CPGNN, and the proposed GCN-SA.
All methods share the same training, validation, and test
splits (48\%, 32\%, 20\% per class).
}
\vspace{-0.3cm}
\scriptsize
\label{all}
\begin{center}
\setlength\tabcolsep{1.7pt}
\begin{tabular}{l c c c c c c c c }
\toprule
Method & Cora & Cites. & Pubmed &  Chame. &Squir. & Corne. & Texas & Wiscon. \\ \hline  
MLP & 74.8$\pm$2.2 & 72.4$\pm$2.2 & 86.7$\pm$0.4 &46.4$\pm$2.5 &29.7$\pm$1.8 &81.1$\pm$6.4 &81.9$\pm$4.8 &85.3$\pm$3.6 \\
GAT & 87.7$\pm$1.9 &75.5$\pm$1.7 &86.7$\pm$1.7 &54.7$\pm$1.9 &30.6$\pm$2.1 &58.9$\pm$3.3 &58.4$\pm$4.5 &55.3$\pm$8.7 \\
GCN & 87.3$\pm$1.3 &76.7$\pm$1.6 &87.4$\pm$0.7&59.8$\pm$2.6 &36.9$\pm$1.4 &57.0$\pm$4.7 &59.5$\pm$5.3 &59.8$\pm$6.9 \\
MixHop&87.6$\pm$0.9 & 76.3$\pm$1.3 & 85.3$\pm$0.6 & 60.5$\pm$2.5 &43.8$\pm$1.5 &73.5$\pm$6.3 &77.8$\pm$5.7 &75.9$\pm$4.9 \\
Geom-GCN &84.9$\pm$1.2 &77.0$\pm$1.4&88.9$\pm$0.9 &60.9$\pm$2.4 &36.1$\pm$1.3&60.8$\pm$6.1 &67.6$\pm$5.8 &64.1$\pm$7.3 \\
IDGL & 88.7$\pm$0.8 & 76.7$\pm$1.2 & 89.4$\pm$0.4 &60.5$\pm$1.9 &42.6$\pm$2.9 &84.5$\pm$4.4 &84.9$\pm$4.1 &87.2$\pm$5.5 \\
H$_{2}$GCN-1&86.9$\pm$1.4&77.1$\pm$1.6 &89.4$\pm$0.3 &57.1$\pm$1.6 &36.4$\pm$1.9 &82.2$\pm$4.8 &84.9$\pm$3.8 &86.7$\pm$4.7 \\
H$_{2}$GCN-2&87.8$\pm$1.4&76.9$\pm$1.8 &89.6$\pm$0.3 &59.4$\pm$2.0 &37.9$\pm$2.1 &82.2$\pm$5.0 &82.2$\pm$4.3 &85.9$\pm$4.2 \\ 
CPGNN& 87.0$\pm$0.8 &76.4$\pm$1.2 &89.6$\pm$0.2&61.0$\pm$3.1& 42.5$\pm$3.1&74.1$\pm$4.9 &75.9$\pm$3.5& 80.6$\pm$5.2\\
\bf{GCN-SA}&\bf{89.4$\pm$1.1}&\bf{77.2$\pm$1.3}&\bf{90.2$\pm$0.5}&\bf{62.5$\pm$2.1}&bf{45.1$\pm$1.0}&\bf{88.4$\pm$6.7}&\bf{89.5$\pm$3.5}&\bf{89.6$\pm$4.0}\\
\bottomrule
\end{tabular}
\end{center}
\end{table*}

\begin{table*}[t]
\caption{Classification accuracies (\%) of the MLP, GAT, GCN, MixHop, Geom-GCN, IDGL, H$_{2}$GCN, CPGNN and the proposed GCN-SA.
All methods share the same training, validation, and test splits (60\%, 20\%, 20\% per class).
} 
\vspace{-3mm}
\scriptsize
\label{all-SEC}
\begin{center}
\setlength\tabcolsep{1.7pt}
\begin{tabular}{l c c c c c c c c }
\toprule
Method & Cora & Cites. & Pubmed &  Chame. &Squir. & Corne. & Texas & Wiscon. \\ \hline  
MLP &76.3$\pm$1.8 &71.8$\pm$2.0 & 87.3$\pm$0.3 & 50.7$\pm$2.1 & 31.1$\pm$1.6 &84.5$\pm$5.2 & 82.1$\pm$5.5 &86.6$\pm$4.3\\
GAT &87.8$\pm$1.7 &76.2$\pm$1.5 & 87.7$\pm$1.5 & 60.6$\pm$1.7 & 36.6$\pm$2.0 &61.3$\pm$3.0 & 63.7$\pm$4.4 &58.3$\pm$7.8\\
GCN &88.2$\pm$1.1 &77.6$\pm$1.4 & 87.1$\pm$0.8 & 61.3$\pm$2.5 & 43.7$\pm$2.2 &60.8$\pm$5.2 & 61.1$\pm$5.7 &62.3$\pm$6.4\\
MixHop   & 88.0$\pm$0.7& 77.1$\pm$1.2 & 86.9$\pm$0.7 & 61.8$\pm$2.3 &44.0$\pm$2.2 & 79.6$\pm$5.9& 81.5$\pm$4.2&80.0$\pm$5.8\\
Geom-GCN & 85.3$\pm$1.0& 78.0$\pm$1.1 &90.1$\pm$0.4 & 60.9$\pm$2.2 &38.1$\pm$2.0&60.8$\pm$5.5&67.6$\pm$5.2&64.1$\pm$6.2\\
IDGL     & 89.1$\pm$0.8& 76.8$\pm$0.9 &89.9$\pm$0.5& 61.9$\pm$1.4 &42.8$\pm$3.0 &86.5$\pm$4.2& 85.9$\pm$4.0&89.1$\pm$6.1\\
H$_{2}$GCN-1  &88.2$\pm$1.1& 77.3$\pm$1.1 &90.0$\pm$0.5&60.2$\pm$2.3 &38.9$\pm$3.0 &84.4$\pm$4.4 &85.5$\pm$4.3 &88.3$\pm$4.2\\
H$_{2}$GCN-2  &88.9$\pm$1.0& 77.6$\pm$1.5 &90.2$\pm$0.5&60.8$\pm$1.8 &40.2$\pm$1.9 &84.0$\pm$4.0 &85.5$\pm$3.8 &87.7$\pm$4.3\\ 
CPGNN &88.0$\pm$0.7& 77.0$\pm$1.0 &88.2$\pm$0.3&62.5$\pm$2.8& 45.3$\pm$2.8&76.5$\pm$5.6 &82.1$\pm$4.2 &82.3$\pm$4.8\\
\bf{GCN-SA}&\bf{90.5$\pm$0.4}&\bf{79.1$\pm$0.6}&\bf{90.9$\pm$0.6}&\bf{64.1$\pm$1.8}&\bf{46.5$\pm$1.6}&\bf{90.8$\pm$4.5}&\bf{89.7$\pm$2.6}&\bf{91.5$\pm$2.5}\\
\bottomrule
\end{tabular}
\end{center}
\end{table*}

\begin{table*}[t]\normalsize
\caption{Hyper-parameter descriptions and range of values tried for the proposed GCN-SA.}
\vspace{-3mm}
\label{hyperparameters}
\begin{center}
\begin{tabular}{c|c|c}
\toprule
Hyper-para.& Descriptions& Range of values tried\\ \hline  %
$lr$  & Learning rate.  & \{0.01, 0.02, \ldots, 0.05\} \\
$d$ & Dropout rate.  & \{0.1, 0.15, \ldots, 0.9\} \\
$wd$ & Weight decay. & \{5e-3, 5e-4, \ldots, 5e-6\} \\
\multirow{2}{*}{$r$} & The number of nearest neighbors r in & \multirow{2}{*}{\{2,3, \ldots, 7\}}\\
&KNN method.&\\
\multirow{2}{*}{$\epsilon$} & Non-negative threshold in similarity & \multirow{2}{*}{\{0.75, 0.8, \ldots, 0.95\} }\\
            &learning.&\\
\multirow{2}{*}{$K$} & The number of rounds in the original & \multirow{2}{*}{\{1, 2, \ldots, 6\}}  \\
            & neighborhood aggregation stage.&\\
\bottomrule
\end{tabular}
\end{center}
\end{table*}

\begin{table*}[th]\vspace{-5mm}
\normalsize
\caption{The hyper-parameters of best accuracy for the proposed GCN-SA on all datasets.}
\label{hyperparameters_value}
\begin{center}
\vspace{-3mm}
\setlength\tabcolsep{1pt}
\begin{tabular}{c|c|c}
\toprule
Dataset& Accu. (\%)&Hyper-parameters\\\hline  %
Cora &  90.5$\pm$0.4 &$lr$:0.03, $wd$:5e-4, $d$:0.55, $\epsilon$:0.95, $K$:6, $seed$:42, $p$:16, $q$:48, $r$:5\\
Citeseer & 79.1$\pm$0.6 &$lr$:0.03, $wd$:5e-4, $d$:0.55, $\epsilon$:0.95, $K$:5, $seed$:42, $p$:16, $q$:48, $r$:2\\
Pubmed & 90.9$\pm$0.6 &$lr$:0.03, $wd$:5e-4, $d$:0.55, $\epsilon$:0.9, $K$:3, $seed$:42, $p$:16, $q$:32, $r$:3\\
Chamele. & 64.1$\pm$1.8&$lr$:0.05, $wd$:5e-4, $d$:0.55,  $\epsilon$:0.85, $K$:1, $seed$:42, $p$:16, $q$:32, $r$:5\\
Squirrel & 46.5$\pm$1.6&$lr$:0.05, $wd$:5e-4, $d$:0.55, $\epsilon$:0.9, $K$:2, $seed$:42, $p$:16, $q$:32, $r$:6\\
Cornell & 90.8$\pm$4.5&$lr$:0.01, $wd$:5e-3,  $d$:0.2, $\epsilon$:0.95, $K$:3, $seed$:42, $p$:16, $q$:32, $r$:3\\
Texas & 89.7$\pm$2.6& $lr$:0.01, $wd$:5e-3, $d$:0.2, $\epsilon$:0.8, $K$:1, $seed$:42, $p$:16, $q$:48, $r$:3\\
Wisconsin &  91.5$\pm$2.5&$lr$:0.01, $wd$:5e-3,  $d$:0.35, $\epsilon$:0.8, $K$:1, $seed$:42, $p$:16, $q$:32, $r$:6\\
\bottomrule
\end{tabular}
\end{center}
\end{table*}

\subsection{Analysis of Time Complexity}\label{ Complexity}

In this paper, $n$ denotes the number of nodes, while $d$ and $q$ represent the number of original and hidden features of nodes, respectively.
$c$ is the class number of node labels.
Given these definitions, the computational complexity of a 2-layer GCN, as shown in \ref{eq:2layerGCN}, is $\mathcal{O}(nq(d+c))$.
Compared with the 2-layer GCN, our GCN-SA contains three additional SA mechanisms and one Hadamard product.
The computational complexities of the first and the second SA mechanisms are the same and equal to $\mathcal{O}(n^{2}q)$.
The computational complexity of the third SA mechanism is equal to $\mathcal{O}(n^{2}c)$.
The Hadamard product's computational complexity is $\mathcal{O}(8nq)$.
In summary, the computational complexity of our GCN-SA is $\mathcal{O}(nq(d+c+8)+2n^{2}q+n^{2}c)$.

In addition, we compare the real running time (500 epochs) of GCN, GAT, Geom-GCN, IDGL, CPGNN, and GCN-SA with the hyper-parameters described in Section \ref{Experimental Setup}.
The Cornell, Texas, and Wisconsin networks possess a similar number of nodes, edges, and node feature dimensions.
Thus, we choose only one dataset, Texas, from the three for the sake of clarity and conciseness.
Figure~\ref{time} exhibits the running time of classification approaches on Texas, Citeseer, Cora, Pubmed, Chameleon, and Squirrel networks. 
As observed, GCN is the fastest.
Due to the time-consuming iterative graph structure learning module, IDGL is the slowest. GAT and GCN-SA are at comparable levels.

\subsection{Comparison Among Different GNNs}\label{cadg}
This experiment estimates the classification performance of the proposed GCN-SA and other outstanding approaches in the literature.
In Table~\ref{all}, the nodes of each class are randomly split into 48\%, 32\%, and 20\% for training, validation, and testing.
In Table~\ref{all-SEC}, the proportions are 60\%, 20\%, and 20\%.
As seen from the Tables~\ref{all},
some GNNs perform even worse than MLP on WebKB networks (low homophily), notably GAT, GCN, MixHop, and Geom-GCN, which have an accuracy of 58.9\%, 57.0\%, 73.5\%, and 60.8\% in Cornell, respectively.
The proposed GCN-SA performs best and achieves 88.4\% in Cornell.
Meanwhile, on graphs with high homophily, our GCN-SA still has advantages.
Compared with the H$_{2}$GCN, IDGL, and CPGNN, GCN-SA has 1.6\%, 0.7\%, and 2.4\% higher accuracies on the Cora dataset.
As can be seen, the proposed GCN-SA possesses the best classification performance.
In addition, similar observations can be made in Table~\ref{all-SEC}.

\section{Conclusion}\label{Conclusion}

In the real world, graphs exhibit varying levels of homophily.
However, most existing GNNs are designed for graphs with high homophily, often underperforming for graphs under heterophily due to their strong homophily assumption. To bridge this gap, we introduce GCN-SA, a novel approach that enhances node embedding through the self-attention mechanism.
During the implementation process of GCN-SA, we present a modified transformer block to incorporate GCN to capture local-to-global correlations.
This integration is pivotal in allowing GCN-SA to handle graphs with varying homophily levels adeptly.
Additionally, we develop a graph structure learning technique powered by self-attention, further refining the GCN-SA's adaptability to diverse graphs.
Moreover, GCN-SA uniquely concatenates multiple node-level embeddings, leveraging their respective strengths to optimize performance. 
The GCN-SA culminates in a versatile and universal node embedding solution in graphs with various homophily levels. 

\subsection{Strengths}\label{strengths}
Here are our work’s main Strengths:
\begin{itemize}
\item[]
$\bullet$ Our GCN-SA is a stable and effective node embedding method that can be used for heterophilic and homophilic graphs.
\end{itemize}

\begin{itemize}
\item[]
$\bullet$ The developed modified transformer block incorporates GCN, allowing GCN to fuse valuable features from the entire graph.
\end{itemize}

\begin{itemize}
\item[]
$\bullet$ The proposed graph-structure-learning strategy is flexible, and others can incorporate it into any GNN model to learn new and reliable graph structures.
\end{itemize}

\begin{itemize}
\item[]
$\bullet$ Our GCN-SA demonstrates superior accuracy in graphs under various levels of homophily. Others can employ GCN-SA to learn node embeddings for graphs with various homophily levels. 
\end{itemize}

\subsection{Limitations}\label{Limitations}
Here are our work’s main limitations:
\begin{itemize}
\item[]
$\bullet$ The advantages of our GCN-SA in homophilic graphs are not particularly significant.
\end{itemize}
\begin{itemize}
\item[]
$\bullet$ The number of new neighbors may be unbalanced when the minimum threshold method is used to screen new neighbors.
\end{itemize}
\begin{itemize}
\item[]
$\bullet$ Introducing multiple transformer blocks increases the computational complexity of the model.
\end{itemize}

\subsection{Future work}\label{ Future}
We propose the following directions for future work:
\begin{itemize}
\item[]
$\bullet$ Enhancing the competitiveness of GCN-SA, particularly for graphs with high-level homophily.
\end{itemize}
\begin{itemize}
\item[]
$\bullet$ Improving the graph-structure-learning method to supply reliable and balanced new neighbors for each node.
\end{itemize}
\begin{itemize}
\item[]
$\bullet$ Reducing the computational complexity of GCN-SA and addressing its scalability.
\end{itemize}
\begin{itemize}
\item[]
$\bullet$ Extending GCN-SA to more challenging scenarios, such as large-scale, dynamic, or knowledge graphs.
\end{itemize}
\section{Acknowledgments}\label{Acknowledgments}
 
This study was supported in part by 
Shaanxi Key Research and Development Program under Grant 2018ZDCXL-GY-04-03-02, in part by the Scientific Research Program of the Education Department of Shaanxi Province under Grant 21JK0762, part by the University-Industry Collaborative Education Program of the Ministry of Education of China under Grant 220802313200859, and part by the National Natural Science
Foundation of China under Grant 42001319.

 \bibliographystyle{elsarticle-num} 
 \bibliography{cas-refs}





\end{document}